# Reasoning about Cardinal Directions between Extended Objects[*]

Xiaotong Zhang, Weiming Liu, Sanjiang Li[†], Mingsheng Ying

Centre for Quantum Computation and Intelligent Systems,
Faculty of Engineering and Information Technology,
University of Technology, Sydney, Broadway NSW 2007, Australia

and

Department of Computer Science and Technology,
Tsinghua University, Beijing 100084, China

October 22, 2018

**Abstract**

Direction relations between extended spatial objects are important commonsense knowledge. Recently, Goyal and Egenhofer proposed a formal model, known as Cardinal Direction Calculus (CDC), for representing direction relations between *connected* plane regions. CDC is perhaps the most expressive qualitative calculus for directional information, and has attracted increasing interest from areas such as artificial intelligence, geographical information science, and image retrieval. Given a network of CDC constraints, the consistency problem is deciding if the network is realizable by connected regions in the real plane. This paper provides a cubic algorithm for checking consistency of basic CDC constraint networks, and proves that reasoning with CDC is in general an NP-Complete problem. For a consistent network of basic CDC constraints, our algorithm also returns a '*canonical*' solution in cubic time. This cubic algorithm is also adapted to cope with cardinal directions between possibly disconnected regions, in which case currently the best algorithm is of time complexity $O(n^5)$.

# 1 Introduction

Representing and reasoning with spatial information is of particular importance in areas such as artificial intelligence (AI), geographical information systems

---

(GISs), robotics, computer vision, image retrieval, natural language processing, *etc.* While the numerical quantitative approach prevails in robotics and computer vision, it is widely acknowledged in AI and GIS that the qualitative approach is more attractive (see *e.g.* [4]).

A predominant part of spatial information is represented by relations between spatial objects. In general, spatial relations are classified into three categories: topological, directional, and metric (*e.g.* size, distance, shape, *etc.*). The RCC8 constraint language [26] is the principal topological formalism in AI, and has been extensively investigated by many researchers (see *e.g.* [29, 27, 34, 5, 37, 36, 18, 19, 17]. When restricted to simple plane regions, RCC8 is equivalent to the 9-Intersection Model (9IM) [6], which is a very influential relation model in GIS.

Unlike topological relations, there are several competitive formal models for direction relations [7, 8, 2]. Most of these models approximate a spatial object by a point (*e.g.* its centroid) or a box. This is too crude in real-world applications such as describing directional information between two countries, say, Portugal and Spain. Recently, Goyal and Egenhofer [11, 10] proposed a relation model, known as cardinal direction calculus (CDC), for representing direction relations between connected plane regions. In CDC the reference object is approximated by a box, while the primary object is unaltered. This means, the exact geometry of the primary object could be used in the representation of the direction. This calculus has 218 basic relations, which is quite large when compared with RCC8 and Allen's Interval Algebra [1]. Due to its expressiveness, CDC has attracted increasing interest from areas such as AI [32, 33, 23], GIS [12], database [31], and image retrieval [14].

One basic criterion for evaluating a formal spatial relation model is the proper balance between its representation expressivity and reasoning complexity. While reasoning complexity of the point-based and the box-based model of direction relations have been investigated in depth (see [20] and [2]), there are few works discussing the complexity of reasoning with CDC.

One central reasoning problem with CDC (and any other qualitative calculus) is the *consistency* (or *satisfaction*) problem. Other reasoning problems such as deriving new knowledge from the given information, updating the given knowledge, or finding a minimal representation can be easily transformed into the consistency problem [4]. In particular, given a network of CDC constraints

$$\mathcal{N} = \{v_i \delta_{ij} v_j\}_{i,j=1}^n \quad \text{(each } \delta_{ij} \text{ is a CDC relation)} \tag{1}$$

over $n$ spatial variables $v_1, \cdots, v_n$, the consistency problem is deciding if the network $\mathcal{N}$ is realizable by a set of $n$ *connected* regions in the real plane. The consistency problem over CDC is an open problem. Before this work, we do not know if there are efficient algorithms deciding if a set of CDC constraints are realizable. Even worse, we do not know if this is a decidable problem. Furthermore, given a satisfiable set of CDC constraints, how to construct a realization in the real plane?

This paper is devoted to solving these problems. We first show by examples that local consistency in particular path-consistency is insufficient for deciding the consistency of basic CDC constraints. Then we devise a cubic algorithm for checking if a network of basic CDC constraints is consistent. In case the network is consistent, this algorithm also generates a solution that is canonical in a sense. Moreover, we also show that deciding the consistency of an arbitrary network of CDC constraints is an NP-Complete problem. This implies that reasoning with CDC is decidable.

Some restricted versions of the consistency problem have been discussed in the literature. Cicerone and di Felice [3] discussed the pairwise consistency problem, which decides when a pair of basic CDC relations $(\delta, \delta')$ is consistent. Skiadopoulos and Koubarakis [32] investigated the weak composition problem [5, 18] of CDC, which is closely related to the consistency problem of basic CDC networks involving only three variables.

The CDC algebra is defined over connected regions. A variant of CDC was proposed in [33], where cardinal directions between possibly disconnected regions are defined in the same way. This calculus, termed $CDC_d$ in this paper, contains 511 basic relations. An $O(n^5)$ algorithm was proposed in [33] for checking the consistency of basic constraints in $CDC_d$, but the consistency problem over CDC is still open. Recently, Navarrete *et al.* [23] tried to adapt the approach used in [33] to cope with connected regions, but their approach turns out to be incorrect.

The remainder of this paper proceeds as follows. Section 2 recalls basic notations in qualitative spatial/temporal reasoning and introduces the well-known Interval Algebra (IA) [1] and Rectangle Algebra (RA) — the two-dimensional counterpart of IA. We introduce the CDC algebra in Section 3, where the connections between CDC and RA relations are established in a natural way. Section 4 introduces the notion of canonical solution of a consistent basic CDC network. Section 5 first proposes an intuitive $O(n^4)$ algorithm for consistency checking of basic networks and then improves it to $O(n^3)$. We apply our main algorithm to the pairwise consistency problem and the weak composition problem over CDC in Section 6. In Section 7 we adapt the main algorithm for connected regions to solve consistency checking in two variants of CDC. Conclusions are given in the last section.

## 2 Qualitative Calculi: Basic Notions and Examples

Since Allen's Interval Algebra, the study of qualitative calculi or relation models has been a central topic in qualitative spatial and temporal reasoning. This section introduces basic notions and important examples of qualitative calculi.

### 2.1 Basic Notions

Let $D$ be a universe of temporal or spatial or spatial-temporal entities. We use small Greek symbols for representing relations on $D$. For a relation $\alpha$ on $D$ and two elements $x, y$ in $D$, we write $(x, y) \in \alpha$ or $x \alpha y$ to indicate that $(x, y)$ is an instance of $\alpha$. For two relations $\alpha, \beta$ on $D$, we define the complement of $\alpha$, the intersection, and the union of $\alpha$ and $\beta$ as follows.

$$\begin{aligned} -\alpha &= \{(x,y) \in D \times D : (x,y) \notin \alpha\} \\ \alpha \cap \beta &= \{(x,y) \in D \times D : (x,y) \in \alpha \text{ and } (x,y) \in \beta\} \\ \alpha \cup \beta &= \{(x,y) \in D \times D : (x,y) \in \alpha \text{ or } (x,y) \in \beta\}. \end{aligned}$$

We write $\mathbf{Rel}(D)$ for the set of binary relations on $D$. Clearly, the 6-tuple $(\mathbf{Rel}(D); -, \cap, \cup, \varnothing, D \times D)$ is a Boolean algebra, where $\varnothing$ and $D \times D$ are, respectively, the empty relation and the universal relation on $D$.

A finite set $\mathcal{B}$ of nonempty relations on $D$ is *jointly exhaustive and pairwise disjoint* (JEPD) if any two entities in $D$ are related by one and only one relation in $\mathcal{B}$. We write $\langle \mathcal{B} \rangle$ for the subalgebra of $\mathbf{Rel}(D)$ generated by $\mathcal{B}$, i.e. the smallest subalgebra of the Boolean algebra $\mathbf{Rel}(D)$ which contains $\mathcal{B}$. Clearly, relations in $\mathcal{B}$ are atoms in the Boolean algebra $\langle \mathcal{B} \rangle$. We call $\langle \mathcal{B} \rangle$ a *qualitative calculus* on $D$, and call relations in $\mathcal{B}$ *basic* relations of the calculus. A similar definition was given by Ligozat and Renz [21], where $\mathcal{B}$ was required to be closed under converse and contain $id_D$ — the identity relation on $D$.

For two relations $\alpha, \beta$ on $D$, the converse of $\alpha$ and the composition of $\alpha$ and $\beta$ are defined as usual.

$$\begin{aligned} \alpha^\smile &= \{(y,x) \in D \times D : (x,y) \in \alpha\} \\ \alpha \circ \beta &= \{(x,y) \in D \times D : (\exists z \in D) \, [(x,z) \in \alpha \text{ and } (z,y) \in \beta]\}. \end{aligned}$$

The composition of two relations $\alpha, \beta$ in $\langle \mathcal{B} \rangle$ is not necessarily in $\langle \mathcal{B} \rangle$, i.e. $\alpha \circ \beta$ cannot be represented as the union of some relations in $\mathcal{B}$. We say a qualitative calculus $\langle \mathcal{B} \rangle$ is *closed under composition* if the composition of any two relations in $\langle \mathcal{B} \rangle$ is still a relation in $\langle \mathcal{B} \rangle$. In general, for $\alpha, \beta \in \langle \mathcal{B} \rangle$, the *weak composition* [5, 18, 28] of $\alpha$ and $\beta$, written as $\alpha \circ_w \beta$, is defined to be the smallest relation in $\langle \mathcal{B} \rangle$ which contains $\alpha \circ \beta$. Clearly, a qualitative calculus is closed under composition if and only if the weak composition operation is the same as the composition operation.

An important reasoning problem in a qualitative calculus $\langle \mathcal{B} \rangle$ is the consistency (or satisfaction) problem. Let $\mathcal{A}$ be a subset of $\langle \mathcal{B} \rangle$. A constraint over $\mathcal{A}$ has the form $(x\gamma y)$ with $\gamma \in \mathcal{A}$. For a set of variables $V = \{v_i\}_{i=1}^n$, and a set of constraints $\mathcal{N}$ involving variables in $V$, we say $\mathcal{N}$ is a *constraint network* if for each pair $(i,j)$ there exists a unique constraint $(v_i \gamma v_j)$ in $\mathcal{N}$. A network $\mathcal{N}$ is said to be over $\mathcal{A}$ if each constraint in $\mathcal{N}$ is over $\mathcal{A}$. In particular, we say a network is a basic network if it is over $\mathcal{B}$. A constraint network $\mathcal{N} = \{v_i \gamma_{ij} v_j\}_{i,j=1}^n$ is *consistent* (or *satisfiable*) if there is an instantiation $\{a_i\}_{i=1}^n$ in $D$ such that $(a_i, a_j) \in \gamma_{ij}$ holds for all $1 \leq i, j \leq n$. In this case, we call $\{a_i\}_{i=1}^n$ a solution of $\mathcal{N}$. The consistency problem over $\mathcal{A}$ is the decision problem of the consistency of constraint networks over $\mathcal{A}$.

## 2.2 Interval Algebra

Interval Algebra (IA) [1] is a qualitative calculus defined on the set of (closed) intervals in the real line. IA is generated by a set $\mathcal{B}_{int}$ of 13 JEPD relations between intervals (see Table 1).

| Relation | Symbol | Converse | Meaning |
|---|---|---|---|
| before | p | pi | $x^- < x^+ < y^- < y^+$ |
| meets | m | mi | $x^- < x^+ = y^- < y^+$ |
| overlaps | o | oi | $x^- < y^- < x^+ < y^+$ |
| starts | s | si | $x^- = y^- < x^+ < y^+$ |
| during | d | di | $y^- < x^- < x^+ < y^+$ |
| finishes | f | fi | $y^- < x^- < x^+ = y^+$ |
| equals | eq | eq | $x^- = y^- < x^+ = y^+$ |

Table 1: Basic IA relations and their converse, where $x = [x^-, x^+], y = [y^-, y^+]$ are two intervals.

IA is closed under converse and composition. This means IA is a relation algebra in the sense of Tarski [35]. The computational complexity of reasoning with IA has been extensively investigated by researchers in artificial intelligence (see [25, 15] and references therein). In particular, Allen [1] introduced the important notion of path-consistency for networks of IA constraints and proved that path-consistency suffices to decide the consistency of basic IA networks. Nebel and Bürckert [25] found a maximal tractable subclass of IA, which is known as *ORD-Horn* and usually denoted by $\mathcal{H}$. For a network of IA constraints over $\mathcal{H}$, path-consistency also suffices to decide consistency.

The definitions of basic IA relations as given in Table 1 concern only the ordering of the endpoints of intervals. This suggests that different solutions of the same basic IA network respect the same ordering. In particular, we could choose intervals that have integer endpoints.

**Definition 1** (canonical set of intervals [22]). Suppose $\mathfrak{m} = \{[m_i^-, m_i^+]\}_{i=1}^n$ is a set of intervals. Set $E(\mathfrak{m})$ to be the set of endpoints of intervals in $\mathfrak{m}$. We say $\mathfrak{m}$ is a *canonical set of intervals* iff $E(\mathfrak{m}) = [0, M] \cap \mathbb{Z}$, where $M$ is the largest number in $E(\mathfrak{m})$. A solution of a basic IA network is called a *canonical interval solution* if it is a canonical set of intervals.

Clearly, if $\mathfrak{m} = \{[m_i^-, m_i^+]\}_{i=1}^n$ is a canonical set of intervals, then each $m_i^-$ ($m_i^+$) is an integer between 0 and $2n - 1$. Moreover, set $M$ to be the largest number in $E(\mathfrak{m})$. Then $M < 2n$ and for any $0 \leq m \leq M$ there exists $i$ such that $m_i^- = m$ or $m_i^+ = m$. The following theorem shows that each consistent basic IA network has a unique canonical solution.

**Theorem 1.** *Suppose $\mathcal{N} = \{v_i \lambda_{ij} v_j\}_{i,j=1}^n$ is a basic IA network. If $\mathcal{N}$ is consistent, then it has a unique canonical interval solution.*

*Proof.* Suppose $\{l_i\}_{i=1}^n$ is a solution of $\mathcal{N}$, where $l_i = [l_i^-, l_i^+]$. Write $\alpha_0 < \alpha_1 < \cdots < \alpha_{n^*}$ for the ordering of $\{l_i^-, l_i^+\}_{i=1}^n$. Define $f, g : \{1, \cdots, n\} \to$

$\{0, 1, \cdots, n^*\}$ as $f(i) = k$ if $l_i^- = \alpha_k$ and $g(i) = k$ if $l_i^+ = \alpha_k$. Let $h_i = [f(i), g(i)]$. Because only the ordering of endpoints of intervals matters in a solution, it is easy to see that $\mathfrak{h} = \{h_i\}_{i=1}^n$ is a solution of $\mathcal{N}$. Set $E(\mathfrak{h})$ to be the set of endpoints of all intervals $h_i$. It is easy to see that $n^* = \max E(\mathfrak{h})$. Moreover, there exists $i$ such that $m = f(i)$ or $m = g(i)$ for each $0 \leq m \leq n^*$. In other words, $E(\mathfrak{h}) = [0, n^*] \cap \mathbb{Z}$. Therefore, $\mathfrak{h}$ is a canonical interval solution of $\mathcal{N}$. Such a solution is clearly unique. $\square$

## 2.3 Rectangle Algebra

Rectangle Algebra (RA) [13, 2] is a qualitative calculus defined on the set of all rectangles in the plane, where we assume a predefined orthogonal basis in the plane and only consider rectangles or boxes two sides of which parallel to the axes of the orthogonal basis.

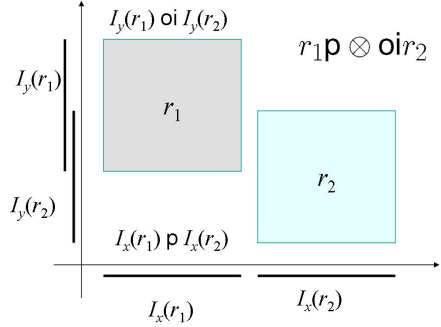

Figure 1: Illustration of Rectangle Relation

For a rectangle $r$, write $I_x(r)$ and $I_y(r)$ as, resp., the $x$- and $y$-projection of $r$. The basic rectangle relation between two rectangles $r_1, r_2$ is defined by the basic IA relation between $I_x(r_1)$ and $I_x(r_2)$ and that between $I_y(r_1)$ and $I_y(r_2)$ (see Fig. 1). More precisely, if $(I_x(r_1), I_x(r_2)) \in \alpha$ and $(I_y(r_1), I_y(r_2)) \in \beta$, then we write $\alpha \otimes \beta$ for the basic rectangle relation between $r_1$ and $r_2$. In other words, for any basic IA relations $\alpha, \beta$,

$$(r_1, r_2) \in \alpha \otimes \beta \Leftrightarrow (I_x(r_1), I_x(r_2)) \in \alpha \text{ and } (I_y(r_1), I_y(r_2)) \in \beta. \qquad (2)$$

Write $\mathcal{B}_{rec}$ for the set of these rectangle relations, *i.e.*

$$\mathcal{B}_{rec} = \{\alpha \otimes \beta : \alpha, \beta \in \mathcal{B}_{int}\} \qquad (3)$$

RA is defined as the qualitative calculus generated by $\mathcal{B}_{rec}$ on the set of rectangles.

**Proposition 1.** *A basic RA network $\mathcal{N} = \{v_i \alpha_{ij} \otimes \beta_{ij} v_j\}_{i,j=1}^n$ ($\alpha_{ij}, \beta_{ij} \in \mathcal{B}_{int}$) is consistent iff its component IA networks $\mathcal{N}_x = \{v_i \alpha_{ij} v_j\}_{i,j=1}^n$ and $\mathcal{N}_y = \{v_i \beta_{ij} v_j\}_{i,j=1}^n$ are consistent.*

*Proof.* This naturally follows from the definition of basic RA relations. □

The definition of canonical solution can be easily extended to RA.

**Definition 2** (canonical rectangle solution). For a consistent basic RA network $\mathcal{N} = \{v_i \alpha_{ij} \otimes \beta_{ij} v_j\}_{i,j=1}^n$ ($\alpha_{ij}, \beta_{ij} \in \mathcal{B}_{int}$), a set of rectangles $\{m_i\}_{i=1}^n$ is a *canonical rectangle solution* of $\mathcal{N}$ iff its $x$- and $y$-projections, $\{I_x(m_i)\}_{i=1}^n$ and $\{I_y(m_i)\}_{i=1}^n$, are canonical interval solutions of $\mathcal{N}_x = \{v_i \alpha_{ij} v_j\}_{i,j=1}^n$ and $\mathcal{N}_y = \{v_i \beta_{ij} v_j\}_{i,j=1}^n$, respectively.

## 3 Cardinal Direction Calculus

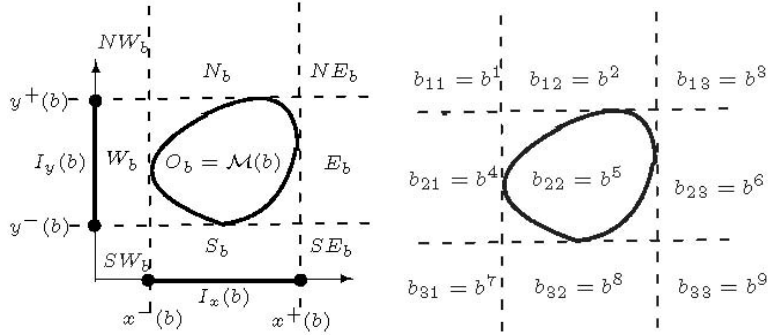

Figure 2: A bounded connected region $b$ and its 9-tiles

In this section we first introduce Cardinal Direction Calculus (CDC) of Goyal and Egenhofer [11] and then establish its connection with Rectangle Algebra. In particular, we will associate a basic RA network $\mathcal{N}_r$ with each basic CDC network $\mathcal{N}$, such that $\mathcal{N}$ is consistent only if $\mathcal{N}_r$ is consistent. More profound connection will be established in the following section.

### 3.1 Direction Relation Matrix

CDC is a qualitative calculus defined for extended two-dimensional objects in the real plane.

**Definition 3** (plane region and connected region). A subset $a$ of the real plane is called a *region* if $a$ is a nonempty regular closed subset, *i.e.* if $a = \overline{a^\circ}$, where $x^\circ$ and $\overline{x}$ are the (topological) interior, and respectively, the (topological) closure of a subset $x$ of the real plane. A region $a$ is said to be *connected* if it has a connected interior $a^\circ$.

For a bounded set $b$ in the real plane, set

$$x^-(b) = \inf\{x : (x,y) \in b\}, \quad x^+(b) = \sup\{x : (x,y) \in b\}, \qquad (4)$$
$$y^-(b) = \inf\{y : (x,y) \in b\}, \quad y^-(b) = \sup\{y : (x,y) \in b\}. \qquad (5)$$

We write
$$I_x(b) = [x^-(b), x^+(b)], \quad I_y(b) = [y^-(b), y^+(b)]. \tag{6}$$
Set
$$\mathcal{M}(b) = I_x(b) \times I_y(b). \tag{7}$$
We call $\mathcal{M}(b)$ the *minimum bounding rectangle* (mbr) of $b$, and call $I_x(b)$ and $I_y(b)$, respectively, the *x*- and *y*-projection of $b$. Clearly, $\mathcal{M}(b)$ is the smallest rectangle containing $b$.

*Remark* 1. The RA relations can be extended from rectangles to arbitrary bounded regions via their minimum bounding rectangles. The extended RA relation between two bounded regions $a$ and $b$ is defined to be the RA relation between $\mathcal{M}(a)$ and $\mathcal{M}(b)$.

By extending the four edges of $\mathcal{M}(b)$, we partition the plane into nine tiles, denoted as $NW_b, N_b, NE_b, W_b, O_b, E_b, SW_b, S_b, SE_b$ (see Fig. 2 (left)). For ease of representation, we also write in sequence $b_{11}, b_{12}, \cdots, b_{33}$ or $b^1, b^2, \cdots, b^9$ for these tiles (see Fig. 2 (right)). Note that each tile is a region, and the intersection of two tiles is either empty or of dimension lower than two.

Since the partition only concerns the mbr of $b$, the following lemma is clear.

**Lemma 1.** *For two bounded regions $b, c$, if $\mathcal{M}(b) = \mathcal{M}(c)$, then $b^i = c^i$ for $i = 1, 2, \cdots, 9$. In particular, if we set $m = \mathcal{M}(b)$, then $b^i = m^i$ for each $i = 1, 2, \cdots, 9$.*

The following notion of direction relation matrix was first proposed by Goyal and Egenhofer [11] for representing the cardinal direction between connected plane regions.

**Definition 4** (direction relation matrix)**.** Suppose $a, b$ are two bounded connected plane regions. Take $b$ as the reference object, and $a$ as the primary object. The direction of $a$ to $b$ is encoded in a $3 \times 3$ Boolean matrix $\mathsf{dir}(a, b) = [d_{ij}]_{i,j=1}^3$, where $d_{ij} = 1$ if and only if $a^\circ \cap b_{ij} \neq \varnothing$, where $a^\circ$ is the interior of $a$ (see Fig. 3). In this case, we call $\mathsf{dir}(a, b)$ a *direction relation matrix*, or a *valid matrix*.

By Lemma 1, it is clear that the direction of $a$ to $b$ is the same as that of $a$ to $\mathcal{M}(b)$.

**Lemma 2.** *For connected regions $a, b, c$, if $\mathcal{M}(b) = \mathcal{M}(c)$, then $\mathsf{dir}(a, b) = \mathsf{dir}(a, c)$. In particular, $\mathsf{dir}(a, b) = \mathsf{dir}(a, \mathcal{M}(b))$.*

This lemma shows that the shape of the reference object is irrelevant: what matters is its mbr.

In what follows we make no distinction between a valid $3 \times 3$ Boolean matrix and the direction relation it represents. It is easy to see that not all matrices are valid. That a matrix is valid is closely related to the following notion of 4-connectedness.

**Definition 5** (4-connected Boolean matrix)**.** A $m \times n$ Boolean matrix $[d_{ij}]_{1 \leq i \leq m, 1 \leq j \leq n}$ is said to be *4-connected* if for any two nonzero cells $pq$ and $st$, there are a series of $k + 1$ nonzero matrix places $p_0q_0, p_1q_1, \cdots, p_kq_k$ such that

- $pq = p_0q_0$ and $st = p_kq_k$; and

- $p_{i+1}q_{i+1}$ is a 4-neighbor of $p_iq_i$, i.e. $|p_{i+1} - p_i| + |q_{i+1} - q_i| = 1$, for each $i = 0, \cdots, k-1$.

In other words, two cells in a Boolean matrix are 4-neighbors if they are horizontally or vertically adjacent, and a Boolean matrix is 4-connected if any two nonzero cells are connected by a series of 4-neighbors.

**Proposition 2.** *[10] A $3 \times 3$ Boolean matrix is a direction relation matrix if and only if it is nonzero and 4-connected.*

Goyal and Egenhofer identified all together 218 valid matrices. For a pair of bounded connected region $a, b$, it is clear that $(a, b)$ determines a unique direction relation matrix, *viz.* $\mathsf{dir}(a, b)$. Write $\mathcal{B}_{dir}$ for the set of cardinal directions represented by these matrices. This means, $\mathcal{B}_{dir}$ is a JEPD set of relations. Cardinal Direction Calculus (CDC) is defined to be the qualitative calculus generated by $\mathcal{B}_{dir}$ over the set of bounded connected regions.

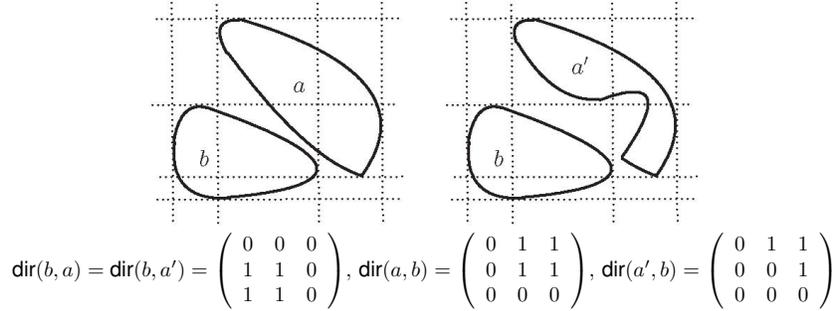

$$\mathsf{dir}(b,a) = \mathsf{dir}(b,a') = \begin{pmatrix} 0 & 0 & 0 \\ 1 & 1 & 0 \\ 1 & 1 & 0 \end{pmatrix}, \mathsf{dir}(a,b) = \begin{pmatrix} 0 & 1 & 1 \\ 0 & 1 & 1 \\ 0 & 0 & 0 \end{pmatrix}, \mathsf{dir}(a',b) = \begin{pmatrix} 0 & 1 & 1 \\ 0 & 0 & 1 \\ 0 & 0 & 0 \end{pmatrix}$$

Figure 3: Illustrations of basic CDC relations

In order to describe the relative position of two connected regions $a, b$, knowing the direction of $b$ to $a$ is not enough. Fig. 3 shows an example, where the direction of $b$ to $a$ is the same as that of $b$ to $a'$ but the direction of $a$ to $b$ is different from that of $a'$ to $b$, i.e. $\mathsf{dir}(b, a) = \mathsf{dir}(b, a')$ but $\mathsf{dir}(a, b) \neq \mathsf{dir}(a', b)$. This is drastically different from IA and many other well-known qualitative calculi. For example, the basic IA relation of $a$ to $b$ is uniquely determined by that of $b$ to $a$. Section 6.1 will investigate this pairwise consistency problem in detail.

In the remainder of this section, we establish the connection between CDC and RA relations.

### 3.2 Direction Relation Vector

We begin with the one-dimensional counterpart of CDC relations.

**Definition 6** (direction relation vector). Suppose $I = [x^-, x^+]$ and $J = [y^-, y^+]$ are two intervals. Interval $J$ partitions the real line into three parts $L_1 =$ $(-\infty, y^-]$, $L_2 = (y^-, y^+)$, and $L_3 = [y^+, +\infty)$. The direction of $I$ to $J$ is encoded in a Boolean vector $\mathsf{dir}(I, J) = (d_1, d_2, d_3)$, where $d_i = 0$ if and only if $(x^-, x^+) \cap L_i = \varnothing$. In this case, we call $(d_1, d_2, d_3)$ a *direction relation vector*.

Clearly, a Boolean vector $t = (t_1, t_2, t_3)$ is a direction relation vector if and only if there exist two intervals $I, J$ such that $t = \mathsf{dir}(I, J)$. The following lemma gives a characterization of direction relation vectors.

**Lemma 3.** *A Boolean vector $t = (t_1, t_2, t_3)$ is a direction relation vector if and only if $t \neq (0, 0, 0)$ and $t \neq (1, 0, 1)$.*

Interestingly, each direction relation vector actually represents an IA relation.

**Lemma 4.** *For two intervals $I, J$, suppose $t = (t_1, t_2, t_3)$ is the direction relation vector of $I$ to $J$. Then we have*

- $t = (1, 0, 0)$ iff $I\mathsf{p}J$ or $I\mathsf{m}J$;
- $t = (0, 1, 0)$ iff $I\mathsf{s}J$ or $I\mathsf{d}J$ or $I\mathsf{f}J$ or $I\mathsf{eq}J$;
- $t = (0, 0, 1)$ iff $I\mathsf{pi}J$ or $I\mathsf{mi}J$;
- $t = (1, 1, 0)$ iff $I\mathsf{o}J$ or $I\mathsf{fi}J$;
- $t = (0, 1, 1)$ iff $I\mathsf{oi}J$ or $I\mathsf{si}J$;
- $t = (1, 1, 1)$ iff $I\mathsf{di}J$.

*Proof.* Take $t = (1, 0, 0)$ as example. For two intervals $I = [x^-, x^+]$ and $J = [y^-, y^+]$, $\mathsf{dir}(I, J) = (1, 0, 0)$ if and only if $x^- < y^-$, $(x^-, x^+) \cap (y^-, y^+) = \varnothing$, and $x^+ < y^+$ hold. This is equivalent to saying that $x^+ \leq y^-$, which is possible if and only if $I\mathsf{p}J$ or $I\mathsf{m}J$. □

In what follows, we call an IA relation a *vector IA relation* if it is the IA relation represented by a direction relation vector. We make no difference between a direction relation vector and the vector IA relation it represents. By the above lemma, we know there are six vector IA relations, *viz.*

$$\mathsf{p} \cup \mathsf{m}, \mathsf{s} \cup \mathsf{d} \cup \mathsf{f} \cup \mathsf{eq}, \mathsf{pi} \cup \mathsf{mi}, \mathsf{o} \cup \mathsf{fi}, \mathsf{oi} \cup \mathsf{si}, \mathsf{di}$$

Note that a vector IA relation is in general non-basic, but a pair of vector IA relations are more precise. For example, from $\mathsf{dir}(I, J) = (0, 1, 0)$, we are not sure whether $I\mathsf{s}J$, or $I\mathsf{d}J$, or $I\mathsf{f}J$, or $I\mathsf{eq}J$ hold. Assuming $\mathsf{dir}(J, I)$ is also given, then it is easy to see that the IA relation between $I, J$ is definite, *i.e.* a basic IA relation.

The following lemma summaries the correspondence between pairs of direction relation vectors and IA relations.

**Lemma 5.** *For a pair of direction relation vectors $(s, t)$ and two intervals $I, J$, we have $s = \mathsf{dir}(I, J)$ and $t = \mathsf{dir}(J, I)$ if and only if $(I, J)$ is an instance of one basic IA relation in the cell specified by $(s, t)$ in Table 2.*

Lemma 5 shows that all basic IA relations except 'meets' and 'before' (and their converses) can be represented as pairs of direction relation vectors.

| $s \setminus t$ | (1,0,0) | (0,1,0) | (0,0,1) | (1,1,0) | (0,1,1) | (1,1,1) |
|---|---|---|---|---|---|---|
| (1,0,0) | ∅ | ∅ | p, m | ∅ | ∅ | ∅ |
| (0,1,0) | ∅ | eq | ∅ | f | s | d |
| (0,0,1) | pi, mi | ∅ | ∅ | ∅ | ∅ | ∅ |
| (1,1,0) | ∅ | fi | ∅ | ∅ | o | ∅ |
| (0,1,1) | ∅ | si | ∅ | oi | ∅ | ∅ |
| (1,1,1) | ∅ | di | ∅ | ∅ | ∅ | ∅ |

Table 2: Pairs of vector IA relations.

### 3.3 Projective IA Networks

The connection between CDC and RA relations is established via the notion of projective interval relations.

**Definition 7** (projective interval relation). For a basic CDC relation $\delta$, the *x-projective interval relation* of $\delta$ is defined as

$$\iota^x(\delta) \equiv \{(I_x(a), I_x(b)) : (a,b) \in \delta\}, \tag{8}$$

where $I_x(a)$ and $I_x(b)$ are the x-projective intervals of $\mathcal{M}(a)$ and $\mathcal{M}(b)$, respectively.

The following lemma proves that each $\iota^x(\delta)$ is indeed a vector IA relation.

**Lemma 6.** *Suppose $\delta = [d_{ij}]_{i,j=1}^3$ is a basic CDC relation. Then the x-projective interval relation $\iota^x(\delta)$ is the IA relation associated to the vector $(d_1, d_2, d_3)$, i.e. $(I, J) \in \iota^x(\delta)$ iff $\text{dir}(I, J) = (d_1, d_2, d_3)$, where $d_j$ is 0 if $\Sigma_{i=1}^3 d_{ij} = 0$ and 1 otherwise.*

*Proof.* See Appendix A. □

Therefore, each x-projective relation $\iota^x(\delta)$ is an IA relation. Immediately, we have

**Lemma 7.** *For a pair of basic CDC relation $(\delta, \gamma)$, if $(a,b)$ is an instance of $\{v_1\delta v_2, v_2\gamma v_1\}$, then $(I_x(a), I_x(b))$ is an instance of $\iota^x(\delta)$ and the converse of $\iota^x(\gamma)$, i.e. $(I_x(a), I_x(b)) \in \iota^x(\delta) \cap \iota^x(\gamma)^\sim$.*

*Proof.* From $(a,b) \in \delta$ and $(b,a) \in \gamma$ we have $(I_x(a), I_x(b)) \in \iota^x(\delta)$ and $(I_x(b), I_x(a)) \in \iota^x(\gamma)$. Therefore, $(I_x(a), I_x(b)) \in \iota^x(\delta) \cap \iota^x(\gamma)^\sim$. □

As a consequence, we have

**Lemma 8.** *A basic CDC constraint network $\mathcal{N} = \{v_i\delta_{ij}v_j\}_{i,j=1}^n$ is consistent only if the IA constraint network $\{v_i\iota_{ij}^x v_j\}_{i,j=1}^n$ is consistent, where $\iota_{ij}^x = \iota^x(\delta_{ij}) \cap \iota^x(\delta_{ji})^\sim$.*

*Proof.* Suppose $\{a_i\}_{i=1}^n$ is a solution to $\mathcal{N}$. Then $\{I_x(a_i)\}_{i=1}^n$ is a solution to $\{v_i\iota_{ij}^x v_j\}_{i,j=1}^n$. □

Note that $\iota_{ij}^x$ is empty or a basic IA relation or the non-basic IA relation $\mathsf{p} \cup \mathsf{m}$ or its converse. Set

$$\mathcal{B}_{int}^* = \{\mathsf{o}, \mathsf{s}, \mathsf{d}, \mathsf{f}, \mathsf{eq}, \mathsf{fi}, \mathsf{di}, \mathsf{si}, \mathsf{oi}\} \cup \{\mathsf{p} \cup \mathsf{m}, \mathsf{pi} \cup \mathsf{mi}\} \qquad (9)$$

For each nonempty IA relation $\iota$ in $\mathcal{B}_{int}^*$, define

$$\widehat{\iota} = \begin{cases} \mathsf{p}, & \text{if } \iota = \mathsf{p} \cup \mathsf{m}; \\ \mathsf{pi}, & \text{if } \iota = \mathsf{pi} \cup \mathsf{mi}; \\ \iota, & \text{otherwise.} \end{cases} \qquad (10)$$

We call $\widehat{\iota}$ the meet-free refinement of $\iota$. Clearly, $\widehat{\iota} = \iota \setminus (\mathsf{m} \cup \mathsf{mi})$ for each $\iota$ in $\mathcal{B}_{int}^*$.

**Lemma 9.** *An IA network $\mathcal{N} = \{v_i \iota_{ij} v_j\}_{i,j=1}^n$ over $\mathcal{B}_{int}^*$ is consistent if and only if $\widehat{\mathcal{N}} = \{v_i \widehat{\iota}_{ij} v_j\}_{i,j=1}^n$ is satisfiable.*

*Proof.* See Appendix B. □

All the above notions and results also apply to the $y$-direction.

**Definition 8** (projective IA networks). For a basic CDC network $\mathcal{N} = \{v_i \delta_{ij} v_j\}_{i,j=1}^n$, recall $\iota_{ij}^x$ is an IA relation defined in Lemma 8. We write $\mathcal{N}_x$ and $\mathcal{N}_y$, resp., for the basic IA networks $\{v_i \rho_{ij}^x v_j\}_{i,j=1}^n$ and $\{v_i \rho_{ij}^y v_j\}_{i,j=1}^n$, where

$$\rho_{ij}^x = \iota_{ij}^x \setminus (\mathsf{m} \cup \mathsf{mi}) \qquad (11)$$
$$\rho_{ij}^y = \iota_{ij}^y \setminus (\mathsf{m} \cup \mathsf{mi}). \qquad (12)$$

We call $\mathcal{N}_x$ and $\mathcal{N}_y$, respectively, the $x$- and $y$-projective IA networks of $\mathcal{N}$.

We write $\mathcal{N}_r$ for the basic RA network $\{v_i \rho_{ij} v_j\}_{i,j=1}^n$, where $\rho_{ij} = \rho_{ij}^x \otimes \rho_{ij}^y$.

As a corollary of Lemma 8 and Lemma 9, we know

**Theorem 2.** *A basic CDC network of constraints $\mathcal{N} = \{v_i \delta_{ij} v_j\}_{i,j=1}^n$ is consistent only if the projective IA networks $\mathcal{N}_x = \{v_i \rho_{ij}^x v_j\}_{i,j=1}^n$ and $\mathcal{N}_y = \{v_i \rho_{ij}^y v_j\}_{i,j=1}^n$ are consistent.*

The above theorem shows that if a basic CDC network is consistent, then its projected IA networks are also consistent. By Prop. 1, this implies that the associated basic RA network $\mathcal{N}_r$ is also consistent.

**Example 1.** *Fig. 4 specifies a basic CDC network $\mathcal{N} = \{v_i \delta_{ij} v_j\}_{i,j=1}^3$, and its associated RA network $\mathcal{N}_r = \{v_i \rho_{ij} v_j\}_{i,j=1}^3$, where $\rho_{ij} = \rho_{ij}^x \otimes \rho_{ij}^y$. For each pair of $i \neq j$, a solution of $\{v_i \delta_{ij} v_j, v_j \delta_{ji} v_i\}$ and a solution of $\{v_i \rho_{ij} v_j\}$ are also illustrated.*

In the next section, we prove that a consistent CDC network $\mathcal{N}$ has a solution $\{a_i\}_{i=1}^n$ such that $\{\mathcal{M}(a_i)\}_{i=1}^n$ is a solution to $\mathcal{N}_r$, the associated basic RA network of $\mathcal{N}$.

| $(i,j)$ | $\delta_{ij}$ | $\delta_{ji}$ | illus. | $\rho_{ij}^x \otimes \rho_{ij}^y$ | illus. |
|---|---|---|---|---|---|
| (1,2) | $\begin{pmatrix} 0 & 1 & 1 \\ 0 & 0 & 1 \\ 0 & 0 & 0 \end{pmatrix}$ | $\begin{pmatrix} 0 & 0 & 0 \\ 1 & 1 & 0 \\ 1 & 1 & 0 \end{pmatrix}$ | | oi $\otimes$ oi | |
| (1,3) | $\begin{pmatrix} 1 & 1 & 0 \\ 0 & 1 & 0 \\ 0 & 0 & 0 \end{pmatrix}$ | $\begin{pmatrix} 0 & 0 & 0 \\ 0 & 0 & 1 \\ 0 & 1 & 1 \end{pmatrix}$ | | o $\otimes$ oi | |
| (2,3) | $\begin{pmatrix} 0 & 0 & 0 \\ 1 & 0 & 0 \\ 0 & 0 & 0 \end{pmatrix}$ | $\begin{pmatrix} 0 & 0 & 1 \\ 0 & 0 & 1 \\ 0 & 0 & 1 \end{pmatrix}$ | | p $\otimes$ d | |

Figure 4: A basic CDC network and its associated basic RA network

### 3.4 Local Consistency Is Insufficient

As we have mentioned in §2.2, path-consistency is sufficient to decide the consistency of a basic IA network. This property is shared by several other qualitative calculi, including RCC8 [24, 29] and the point-based cardinal direction calculus of Ligozat [20]. This property, however, does not hold for CDC.

We first note that a basic network is path-consistent if and only if every subnetwork involving three variables is consistent. The following example shows that there is a basic CDC network that is path-consistent but still inconsistent. Similar conclusion has been obtained in [33] for the case of possibly disconnected regions.

**Example 2.** Let $a_1$, $a_2$, $a_3$, $a_4$ be the squares illustrated in Fig. 5. Consider the basic CDC network $\mathcal{N} = \{v_i \delta_{ij} v_j\}_{i,j=1}^5$, where

- $\delta_{ij} = \mathsf{dir}(a_i, a_j)$ for $1 \leq i, j \leq 4$,

- $\delta_{i5} = \begin{pmatrix} 0 & 0 & 0 \\ 0 & 1 & 0 \\ 0 & 0 & 0 \end{pmatrix}$ for $1 \leq i \leq 4$, and

- $\delta_{51} = \begin{pmatrix} 0 & 1 & 1 \\ 0 & 0 & 1 \\ 0 & 0 & 0 \end{pmatrix}$, $\delta_{52} = \delta_{53} = \begin{pmatrix} 1 & 1 & 1 \\ 1 & 0 & 1 \\ 0 & 0 & 0 \end{pmatrix}$, $\delta_{54} = \begin{pmatrix} 1 & 1 & 0 \\ 1 & 0 & 0 \\ 0 & 0 & 0 \end{pmatrix}$.

Fig. 5 and Fig. 6 show that every subnetwork of $\mathcal{N}$ which involves four variables is satisfiable. This implies in particular that $\mathcal{N}$ is path-consistent.

$\mathcal{N}$ is, however, inconsistent. We prove this by contradiction. Suppose $\{b_i\}_{i=1}^5$ is a solution to $\mathcal{N}$. Then $\mathcal{M}(b_i)(1 \leq i \leq 4)$ will be related in a configuration similar to that of $a_i$ in Fig. 5. By the assumption of $\delta_{i5}$ for $1 \leq i \leq 4$, we know $\mathcal{M}(b_i)$ is contained in $\mathcal{M}(b_5)$ for each $i = 1, 2, 3, 4$. By the definition of $\delta_{51}$, we further conclude that $\mathcal{M}(b_1)$ contains the lower left corner of $\mathcal{M}(b_5)$.

Similarly, $\mathcal{M}(b_4)$ contains the lower right corner of $\mathcal{M}(b_5)$. Meanwhile, according to $\delta_{5i}$ we have $b_5^\circ \cap \mathcal{M}(b_i) = \varnothing$ for $1 \leq i \leq 4$. By the configuration of $b_1, b_2, b_3, b_4$ (c.f. Fig. 5), this means no point of $b_5^\circ$ can be below the line $y = y_0$, where $y_0 = \sup\{y : (x,y) \in \mathcal{M}(b_1)\}$. This is impossible since $\mathcal{M}(b_1) \subset \mathcal{M}(b_5)$. Therefore, $\mathcal{N}$ is inconsistent.

In fact, for any positive integer $k$, the consistency of all subnetworks of $\mathcal{N}$ with $k$ variables does not guarantee the consistency of $\mathcal{N}$. A counter example can be constructed analogously (see Fig. 7, where $k$ squares instead of four are used). This shows that local consistency is insufficient for consistency even for basic CDC networks. To determine if a CDC network is consistent, we need to consider the constraints in a network as a whole.

## 4 Canonical Solution

Suppose $\mathcal{N} = \{v_i \delta_{ij} v_j\}_{i,j=1}^n$ is a consistent basic CDC network, where each $\delta_{ij}$ is in $\mathcal{B}_{dir}$. We show $\mathcal{N}$ has a canonical solution in a sense similar to that for IA network. We begin with an arbitrary solution $\mathfrak{a} = \{a_i\}_{i=1}^n$ of $\mathcal{N}$, then transform regions in $\mathfrak{a}$ step by step into regions in the digital plane $\mathbb{Z}^2$ without changing the cardinal direction relations between any two of these regions.

### 4.1 Regular Solution

Suppose $\mathfrak{a} = \{a_i\}_{i=1}^n$ is a set of $n$ connected regions. This subsection shows how to *regularize* $\mathfrak{a}$. Illustrations are given in Fig. 8.

Write $m_i = [x_i^-, x_i^+] \times [y_i^-, y_i^+]$ for the mbr of $a_i$. Set

$$e^- = \min\{x_i^- : 1 \leq i \leq n\}, \tag{13}$$
$$e^+ = \max\{x_i^+ : 1 \leq i \leq n\}, \tag{14}$$
$$f^- = \min\{y_i^- : 1 \leq i \leq n\}, \tag{15}$$
$$f^+ = \max\{y_i^+ : 1 \leq i \leq n\}. \tag{16}$$

Let

$$S(\mathfrak{a}) = [e^-, e^+] \times [f^-, f^+]. \tag{17}$$

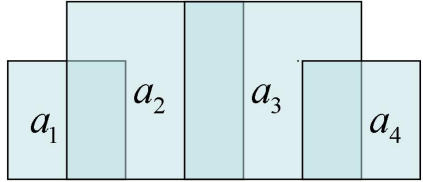

Figure 5: A solution to the subnetwork over $\{v_1, v_2, v_3, v_4\}$ of $\mathcal{N}$ in Example 2

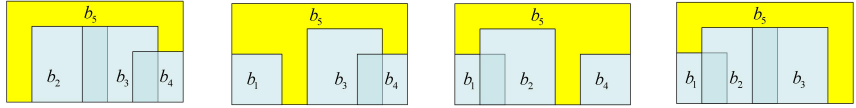

Figure 6: Solutions to other subnetworks of $\mathcal{N}$ in Example exam:path-consistencyNOTcons which involve four variables

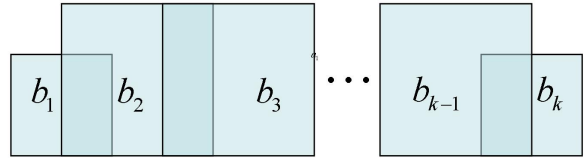

Figure 7: A configuration of $k$ squares

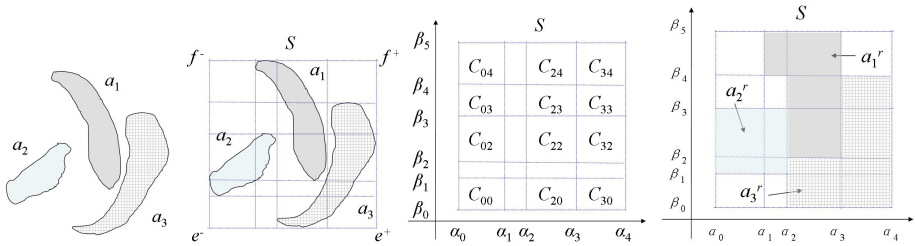

Figure 8: Illustration of regularization

Extending edges of each rectangle $m_i$ until meeting the boundary of $S(\mathfrak{a})$, we partition $S(\mathfrak{a})$ into small cells. Suppose

$$\alpha_0 < \alpha_1 < \cdots < \alpha_{n_x} \tag{18}$$

is the ordering of real numbers in $\{x_i^-, x_i^+ : 1 \leq i \leq n\}$, and

$$\beta_0 < \beta_1 < \cdots < \beta_{n_y} \tag{19}$$

is the ordering of real numbers in $\{y_i^-, y_i^+ : 1 \leq i \leq n\}$. Denote

$$c_{ij} = [\alpha_i, \alpha_{i+1}] \times [\beta_j, \beta_{j+1}], \quad (0 \leq i < n_x, 0 \leq j < n_y) \tag{20}$$

and write

$$C(\mathfrak{a}) = \{c_{ij} : 0 \leq i < n_x, 0 \leq j < n_y\} \tag{21}$$

for the set of these cells. For each $i$, let

$$a_i^r = \bigcup \{c \in C(\mathfrak{a}) : c \cap a_i^\circ \neq \varnothing\}. \tag{22}$$

**Definition 9** (frame, cell set, regularization). For a set of connected regions $\mathfrak{a} = \{a_i\}_{i=1}^n$, we call $S(\mathfrak{a})$ the *frame* of $\mathfrak{a}$, and call $C(\mathfrak{a})$ the *cell set* of $\mathfrak{a}$, and call $\mathfrak{a}^r = \{a_i^r\}_{i=1}^n$ the *regularization* of $\mathfrak{a}$, where $S(\mathfrak{a})$, $C(\mathfrak{a})$, and $a_i^r$ are defined in Equations 17,21,22, respectively.

The regularization $\mathfrak{a}^r$ has the same frame and the same cell sets with $\mathfrak{a}$.

**Lemma 10.** *For a set of connected regions $\mathfrak{a} = \{a_i\}_{i=1}^n$, we have $a_i \subseteq a_i^r$ for each $i$, and $S(\mathfrak{a}) = S(\mathfrak{a}^r)$ and $C(\mathfrak{a}) = C(\mathfrak{a}^r)$, where $\mathfrak{a}^r = \{a_i^r\}_{i=1}^n$ is the regularization of $\mathfrak{a}$.*

*Proof.* Straightforward. □

**Example 1 (continued)**
For the constraint network specified in Fig. 4, Fig. 8 illustrates how to transform a solution $\{a_1, a_2, a_3\}$ (leftmost of Fig. 8) to a regular solution $\{a_1^r, a_2^r, a_3^r\}$ (rightmost of Fig. 8). The frame and cell set are also illustrated.

If $\mathfrak{a}$ is a solution to a basic CDC network $\mathcal{N}$, then so is its regularization.

**Lemma 11.** *Suppose $\mathfrak{a} = \{a_i\}_{i=1}^n$ is a solution to a basic CDC network $\mathcal{N}$. Then $\mathfrak{a}^r = \{a_i^r\}_{i=1}^n$ is also a solution to $\mathcal{N}$.*

*Proof.* It is clear that each $a_i^r$ is connected, and $\mathcal{M}(a_i^r) = \mathcal{M}(a_i)$. It is straightforward to show that $\mathsf{dir}(a_i, a_j) = \mathsf{dir}(a_i^r, a_j^r)$ for any $i, j$. □

**Definition 10** (regular solution). A solution $\mathfrak{a} = \{a_i\}_{i=1}^n$ of a basic CDC network $\mathcal{N}$ is called *regular* if $\mathfrak{a}$ is the same as its regularization $\mathfrak{a}^r = \{a_i^r\}_{i=1}^n$.

It is easy to see that the regularization of a solution is regular. This is because $\mathfrak{a}^r$ and $\mathfrak{a}$ have the same frame and the same cell set. By Eq. 22, it is straightforward to see that a solution $\mathfrak{a} = \{a_i\}_{i=1}^n$ of a basic CDC network $\mathcal{N}$ is regular if and only if each $a_i$ is the union of all cells in $C(\mathfrak{a})$ that has nonempty intersection with the interior of $a_i$.

## 4.2 Meet-Free Solution

We next show each consistent basic CDC network $\mathcal{N}$ has a solution that is meet-free in the following sense.

**Definition 11** (meet-free solution). A solution $\mathfrak{a} = \{a_i\}_{i=1}^n$ of $\mathcal{N}$ is *meet-free* if for any $i, j$, $I_x(a_i)$ does not meet $I_x(a_j)$, and $I_y(a_i)$ does not meet $I_y(a_j)$.

Suppose $\mathfrak{a} = \{a_i\}_{i=1}^n$ is a solution of $\mathcal{N}$ and $\mathfrak{a}^r = \{a_i^r\}_{i=1}^n$ is its regularization (see Fig. 11). Suppose $m_i = [x_i^-, x_i^+] \times [y_i^-, y_i^+]$ and $m_j = [x_j^-, x_j^+] \times [y_j^-, y_j^+]$ meet at $x$ direction, i.e. $x_i^+ = x_j^-$. Recall $\alpha_0 < \cdots < \alpha_{n_x}$ is the ordering of $\{x_i^-, x_i^+ : 1 \leq i \leq n\}$. We call $\alpha_k = x_i^+$ an $x$-meet point. Clearly, $k > 0$ and $x_i^- \leq \alpha_{k-1} < x_i^+ = x_j^- = \alpha_k < \alpha_{k+1} \leq x_j^+$. We next show how to delete this meet point by transforming $\mathfrak{a}^r$ into another regular solution.

Write $\alpha^* = (\alpha_k + \alpha_{k+1})/2$. The line $x = \alpha^*$ divides each cell $c_{kl}$ ($0 \leq l < n_y$) into two equal parts, written in order $c_{kl}^-$ and $c_{kl}^+$. For each $1 \leq s \leq n$ and each $0 \leq l < n_y$, if $c_{kl} \subseteq a_s^r$ but $c_{k-1,l} \not\subseteq a_s^r$ then delete $c_{kl}^-$ from $a_s^r$. The remaining part of each $a_s^r$, written as $b_s$, is still connected, and it is straightforward to show that $\mathfrak{b} = \{b_s\}_{s=1}^n$ is also a regular solution of $\mathcal{N}$. Such a modification introduces no new meet points. Continuing this process for at most $n$ times, we will have a solution that has no $x$-meet points. The same modification can be applied to $y$-meet points. In this way we obtain a meet-free solution. An example is shown in Fig. 9. It is easy to see that the meet-free solution has the same frame as $\mathfrak{a}$. The two solutions, however, have different cell sets.

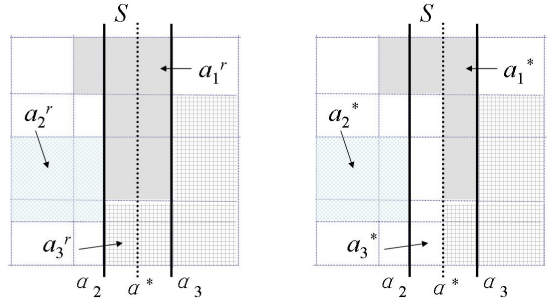

Figure 9: Illustration of meet-freeing

In conclusion, we have the following lemma:

**Lemma 12.** *Each consistent basic CDC network has a regular solution that is meet-free.*

So we can safely assume that $\mathfrak{a}$ is a meet-free solution that is regular, i.e. $\mathfrak{a}^r = \mathfrak{a}$.

## 4.3 Digital Solution and Canonical Solution

Suppose $\mathfrak{a}$ is a meet-free solution of $\mathcal{N}$ that is regular. We next transform $\mathfrak{a}$ into a solution in the digital plane $\mathbb{Z}^2$.

**Definition 12** (pixel, digital region, digital solution). A *pixel* is a rectangle $p_{ij} = [i, i+1] \times [j, j+1]$, where $i, j$ are integers. A region $a$ is *digital* if $a$ is composed of pixels, i.e. $p_{ij} \cap a^\circ \neq \varnothing$ iff $p_{ij} \subseteq a$. A solution $\mathfrak{a} = \{a_i\}_{i=1}^n$ of a basic CDC network is *digital* if each $a_i$ is a digital region.

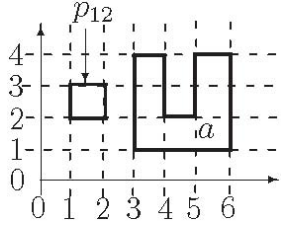

Figure 10: A pixel and a digital region $a$

As in the paragraph immediately above Lemma 11, we write $m_i = [x_i^-, x_i^+] \times [y_i^-, y_i^+]$ for the mbr of $a_i$. Recall $S(\mathfrak{a})$ and $C(\mathfrak{a}) = \{c_{ij}\}_{0 \leq i < n_x, 0 \leq j < n_y}$ are, respectively, the frame and the cell set of $\mathfrak{a}$. Since $\mathfrak{a}$ is regular, each $a_i$ is composed of a subset of cells in $C(\mathfrak{a})$.

We next show how to transform the solution $\mathfrak{a} = \{a_s\}_{s=1}^n$ into a digital solution $\mathfrak{b} = \{b_s\}_{s=1}^n$. For $1 \leq s \leq n$, define a subset $b_s$ of $S(\mathfrak{a}) = [0, n_x] \times [0, n_y]$ as follows:

A pixel $p_{ij}$ is contained in $b_s$ iff the cell $c_{ij}$ in $C(\mathfrak{a})$ is contained in $a_s$.

That is,
$$b_s = \bigcup \{p_{ij} : c_{ij} \subseteq a_s, 0 \leq i < n_x, 0 \leq j < n_y\} \qquad (23)$$

Clearly, $b_s$ is a connected digital region (See Fig. 11 for illustration). By definition we have $\mathsf{dir}(a_i, a_j) = \mathsf{dir}(b_i, b_j)$ for any $i, j$. Therefore, the assignment $\mathfrak{b} = \{b_s\}_{s=1}^n$ is also a solution of $\mathcal{N}$. We observe that $\mathfrak{b}$ has several good properties.

**Lemma 13.** *Suppose $\mathfrak{a} = \{a_s\}_{s=1}^n$ is a regular meet-free solution of a basic CDC network $\mathcal{N}$. The digital solution $\mathfrak{b} = \{b_s\}_{s=1}^n$ constructed as above is also regular and meet-free. Moreover, $\{I_x(b_s)\}_{s=1}^n$ and $\{I_y(b_s)\}_{s=1}^n$ are the canonical solutions of the projective IA networks $\mathcal{N}_x$ and $\mathcal{N}_y$, respectively.*

*Proof.* That $\mathfrak{b}$ is regular and meet-free follows directly from the properties of $\mathfrak{a}$. We show $\{I_x(b_s)\}_{s=1}^n$ and $\{I_y(b_s)\}_{s=1}^n$ are canonical sets of intervals. Take the $x$-direction as example. Because $\mathfrak{b} = \{b_i\}_{i=1}^n$ is a solution to $\mathcal{N}$, we know $(I_x(b_i), I_x(b_j))$ is an instance of $\iota^x(\delta_{ij}) \cap \iota^x(\delta_{ji})^\sim$. Since $\mathfrak{b}$ is meet-free, $I_x(b_i)$ cannot meet $I_x(b_j)$ for any two $i, j$. This implies that $\{I_x(b_i)\}_{i=1}^n$ is a solution to the basic IA network $\mathcal{N}_x = \{v_i \rho^x_{ij} v_j\}^n_{i,j=1}$, where $\rho^x_{ij}$ is defined by Eq. 11. By the choice of $\alpha_i$ (Eq. 18), we can easily show $\{I_x(b_i)\}^n_{i=1}$ is a canonical set of intervals. $\square$

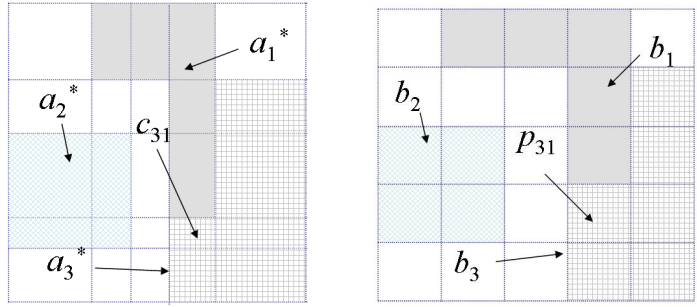

Figure 11: Transform a regular solution into a digital one

We call the solution $\mathfrak{b}$ in the above lemma a canonical solution. More precisely, we have

**Definition 13** (canonical solution). *A solution $\mathfrak{b} = \{b_i\}^n_{i=1}$ of $\mathcal{N}$ is said to be canonical if it is digital, regular, and meet-free, and $\{I_x(b_i)\}^n_{i=1}$ and $\{I_y(b_i)\}^n_{i=1}$ are canonical sets of intervals.*

As a corollary of Lemma 13, we have

**Theorem 3.** *Each consistent basic CDC network has a canonical solution.*

Canonical solutions are not necessarily unique. For a consistent basic CDC network $\mathcal{N} = \{v_i \delta_{ij} v_j\}^n_{i,j=1}$, recall we write $\mathcal{N}_x = \{v_i \rho^x_{ij} v_j\}^n_{i,j=1}$ and $\mathcal{N}_y = \{v_i \rho^y_{ij} v_j\}^n_{i,j=1}$ for the two projective basic IA networks (see Dfn. 8). By Thm. 2, we know both $\mathcal{N}_x$ and $\mathcal{N}_y$ are consistent. Suppose $\mathcal{I} = \{I_i\}^n_{i=1}$ and $\mathcal{J} = \{J_i\}^n_{i=1}$ are their canonical interval solutions as given by Thm. 1, where $I_i = [x_i^-, x_i^+]$, $J_i = [y_i^-, y_i^+]$. Write $n_x = \max\{x_i^+\}^n_{i=1} < 2n$, $n_y = \max\{y_i^+\}^n_{i=1} < 2n$. Set $T \equiv [0, n_x] \times [0, n_y]$ and $m_i = I_i \times J_i$. Clearly, $m_i \subseteq T$. By Thm. 3, $\mathcal{N}$ has a canonical solution because of its consistency.

**Lemma 14.** *If $\mathfrak{a} = \{a_i\}^n_{i=1}$ is a canonical solution of $\mathcal{N}$, then each $a_i$ is a connected digital region whose mbr is $m_i = I_i \times J_i$, where $\{I_i\}^n_{i=1}, \{J_i\}^n_{i=1}$ are, respectively, the canonical interval solutions of $\mathcal{N}_x$ and $\mathcal{N}_y$.*

As a corollary, we have

**Corollary 1.** *Suppose $\mathfrak{b} = \{b_i\}^n_{i=1}$ and $\mathfrak{b}' = \{b'_i\}^n_{i=1}$ are two canonical solutions of $\mathcal{N}$. Then $I_x(b_i) = I_x(b'_i)$, $I_y(b_i) = I_y(b'_i)$, and $\mathcal{M}(b_i) = \mathcal{M}(b'_i)$.*

This shows that two canonical solutions have the same mbrs. Moreover, it can be proved that the union of two canonical solutions is also a canonical solution. Is there a maximal canonical solution? Where a canonical solution $\mathfrak{a} = \{a_i\}_{i=1}^n$ of a basic CDC network $\mathcal{N}$ is said to be *maximal* if for any other canonical solution $\mathfrak{a}' = \{a_i'\}_{i=1}^n$ of $\mathcal{N}$ we have $a_i' \subseteq a_i$ for $i = 1, \cdots, n$.

In the next subsection, we give a method for constructing the maximal canonical solution.

### 4.4 Maximal Canonical Solution

For a consistent basic CDC network $\mathcal{N}$, suppose $\mathfrak{a} = \{a_i\}_{i=1}^n$ is a canonical solution of $\mathcal{N}$. By Lemma 14 and the definition of canonical solution, we know $\{I_x(a_i)\}_{i=1}^n$ and $\{I_y(a_i)\}_{i=1}^n$ are the canonical interval solutions of, respectively, $\mathcal{N}_x$ and $\mathcal{N}_y$. Set $I_i = I_x(a_i)$, $J_i = I_y(a_i)$, and $m_i = I_i \times J_i$ for each $i$. Because of the constraints in $\mathcal{N}$, not all pixels in $m_i$ can appear in $a_i$.

Note a basic CDC relation $\delta = [d_{st}]_{s,t=1}^3$ can be rewritten as a 9-tuple $\delta = (d^\phi)_{\phi=1}^9$ by setting $d^\phi = d_{st}$, where $\phi = 3(s-1) + t$. For each variable $v_j$ and each $1 \leq \phi \leq 9$, we write $m_j^\phi$ for the $\phi$-th tile associated to the rectangle $m_j$ (see Fig. 2 (right)). For each constraint $\delta_{ij} = (d_{ij}^\phi)_{\phi=1}^9$ of $v_i$ to $v_j$, by $(a_i, a_j)$ satisfies $\delta_{ij}$ and $\mathcal{M}(a_j) = m_j$, we know $d_{ij}^\phi = 0$ if and only if $a_i^\circ \cap m_j^\phi = \varnothing$. Therefore, each pixel contained in $m_j^\phi$ is disallowed in $a_i$ if $d_{ij}^\phi = 0$. For each $i$, we write

$$D_i = \{p_{st} \subseteq m_i : \text{there exist } j \neq i, \phi \text{ such that } p_{st} \subseteq m_j^\phi \text{ and } d_{ij}^\phi = 0\} \quad (24)$$

That is, $D_i$ contains all pixels in $m_i$ that are disallowed for violating some constraint in $\mathcal{N}$. Set

$$b_i = \bigcup \{p_{st} \subseteq m_i : p_{st} \notin D_i\} \quad (25)$$

**Lemma 15.** *If $\mathfrak{a} = \{a_i\}_{i=1}^n$ is a canonical solution of $\mathcal{N}$, then $\mathcal{M}(a_i) = \mathcal{M}(b_i)$ and $a_i^\circ \cap p_{st} = \varnothing$ for any $p_{st} \in D_i$, i.e. $a_i \subseteq b_i$, where $b_i$ is defined in Eq. 25.*

We note that $b_i$ may be disconnected. As a connected region, $a_i$ must be contained in a unique connected component of $b_i$. We observe that two connected digital regions share at least one pixel in common if they have the same mbr.

**Lemma 16.** *Let $a$ and $b$ be two connected digital regions. If $a^\circ \cap b^\circ = \varnothing$, then $\mathcal{M}(a) \neq \mathcal{M}(b)$.*

*Proof.* Without loss of generality, suppose $\mathcal{M}(a) = [0, n_1] \times [0, n_2]$, where $n_1, n_2$ are positive integers. Suppose $a^\circ \cap b^\circ = \varnothing$ and $\mathcal{M}(a) = \mathcal{M}(b)$. Since $\mathcal{M}(a) = [0, n_1] \times [0, n_2]$, $a$ contains a pixel $p_{0s}$ and a pixel $p_{n_1-1,t}$ for some $s, t \in \{0, 1, \cdots, n_2-1\}$. Because $a$ is a connected digital region, there is a path $\alpha$ (*i.e.* a sequence of pixels in $a$) that connects $p_{0s}$ to $p_{n_1-1,t}$. Clearly, $\alpha$ separates $\mathcal{M}(a)$ into at least two disjoint components, each of which is contained in a rectangle that is smaller than $\mathcal{M}(a)$. Since $a^\circ \cap b^\circ = \varnothing$, we know $b$ is contained in one such component. Therefore, $\mathcal{M}(b)$ is smaller than $\mathcal{M}(a)$. A contradiction. □

By the above lemma, we know each $b_i$ has at most one connected component $c_i$ such that $\mathcal{M}(c_i) = m_i$. Moreover, if $\mathcal{N}$ is consistent, then each $b_i$ has a unique component $c_i$ with $\mathcal{M}(c_i) = \mathcal{M}(b_i) = m_i$.

**Lemma 17.** *Suppose $\mathcal{N} = \{v_i \delta_{ij} v_j\}_{i,j=1}^n$ is a consistent basic CDC network. Then each $b_i$ has a unique connected component $c_i$ such that $\mathcal{M}(c_i) = \mathcal{M}(b_i) = m_i$.*

*Proof.* Suppose $\{a_i\}_{i=1}^n$ is a canonical solution. We have $a_i$ is connected and $\mathcal{M}(a_i) = m_i$. By Lemma 15, we know $a_i \subseteq b_i$. Let $c_i$ be the unique component of $b_i$ which contains $a_i$. Then by $m_i = \mathcal{M}(a_i) \subseteq \mathcal{M}(c_i) \subseteq \mathcal{M}(b_i) = m_i$ we know $\mathcal{M}(c_i) = m_i$. □

Moreover, $\{c_i\}_{i=1}^n$ is also a solution of $\mathcal{N}$.

**Lemma 18.** *Let $c_i$ be the unique connected component of $b_i$ such that $\mathcal{M}(c_i) = m_i$. Then $\mathfrak{c} = \{c_i\}_{i=1}^n$ is a solution of $\mathcal{N}$. Moreover, if $\mathfrak{a} = \{a_i\}_{i=1}^n$ is a canonical solution of $\mathcal{N}$, then $a_i \subseteq c_i$ for each $i$.*

*Proof.* That $a_i$ is contained in $c_i$ is clear. We need only show $\mathsf{dir}(c_i, c_j) = \delta_{ij}$ also holds for each pair of $i, j$. This is equivalent to proving for all $1 \leq \phi \leq 9$ that $c_i^\circ \cap m_j^\phi = \varnothing$ if and only if $d_{ij}^\phi = 0$. Recall $a_i \subseteq c_i \subseteq b_i \subseteq m_i = \mathcal{M}(a_i)$. If $d_{ij}^\phi = 0$, by $c_i \subseteq b_i$, for each pixel $p_{st} \in D_i$, we know $c_i^\circ \cap p_{st} = \varnothing$, hence $c_i^\circ \cap m_j^\phi = \varnothing$. On the other hand, if $d_{ij}^\phi = 1$, then $c_i^\circ \cap m_j^\phi \supseteq a_i^\circ \cap m_j^\phi \neq \varnothing$. This shows $\mathfrak{c}$ is a solution of $\mathcal{N}$. □

It is easy to see that each region $c_i$ in $\mathfrak{c}$ is a connected digital region. Moreover, $\mathfrak{c}$ is a regular and meet-free solution of $\mathcal{N}$ such that $I_x(c_i) = I_x(a_i)$ and $I_y(c_i) = I_y(a_i)$ for each $i$. By Dfn. 13, we know $\mathfrak{c}$ is also a canonical solution of $\mathcal{N}$. By $a_i \subseteq c_i$ and the arbitrariness of $\mathfrak{a}$, we know $\mathfrak{c}$ is the maximal canonical solution of $\mathcal{N}$.

**Theorem 4.** *Suppose $\mathcal{N} = \{v_i \delta_{ij} v_j\}_{i,j=1}^n$ is a consistent basic CDC network. Let $c_i$ be the unique connected component of $b_i$ of Eq. 25 such that $\mathcal{M}(c_i) = m_i$. Then $\mathfrak{c} = \{c_i\}_{i=1}^n$ is the maximal canonical solution of $\mathcal{N}$.*

It is worth stressing that each $m_i$, $b_i$, $c_i$ are independent of the choice of canonical solution $\mathfrak{a}$. They only depend on the specific constraint network $\mathcal{N}$.

An example is given in Fig. 12 (left) to illustrate the procedures. Note that each $b_i$ obtained in this example happens to be connected.

**Example 1 (continued)**
For the network specified in Fig. 4, we have

$$x_2^- < x_1^- < x_2^+ < x_3^- < x_1^+ < x_3^+ \tag{26}$$
$$y_3^- < y_2^- < y_1^- < y_2^+ < y_3^+ < y_1^+ \tag{27}$$

The canonical interval solutions of $\mathcal{N}_x$ and $\mathcal{N}_y$ are illustrated in Fig. 12 (left). Suppose $\{a, b, c\}$ is a solution of the network described in Fig. 4. Note that the (2,2)-entry of $\delta_{12}$ is 0, which is possible only if $a^\circ \cap \mathcal{M}(b) = \varnothing$. This excludes pixel $p_{12}$ from $a$ (see Fig. 12). Fig. 12 (right) illustrates the maximal canonical solution $\{a, b, c\}$ of $\mathcal{N}$.

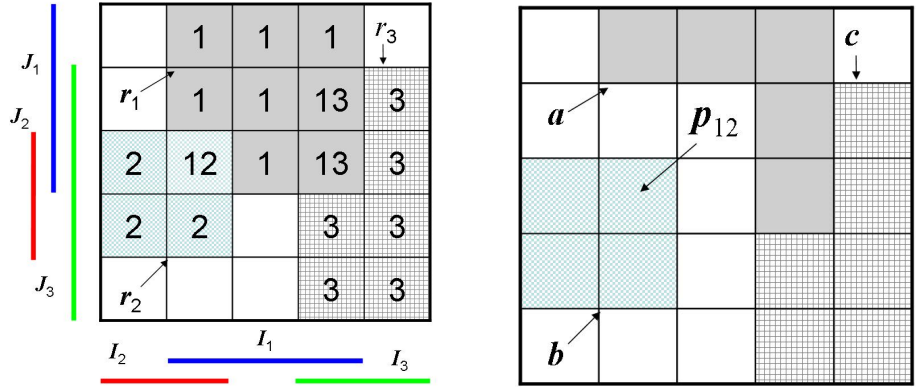

Figure 12: Canonical interval solutions and maximal canonical solution

## 5  A Consistency Checking Algorithm

In this section, we describe our algorithm for checking consistency of basic CDC networks. As in the last section, we assume $\mathcal{N} = \{v_i \delta_{ij} v_j\}_{i,j=1}^n$ is a basic CDC network. To examine the consistency of $\mathcal{N}$, we first compute the two projective IA networks $\mathcal{N}_x$ and $\mathcal{N}_y$. If any of these two networks is inconsistent, then $\mathcal{N}$ is inconsistent, either. Assume both $\mathcal{N}_x$ and $\mathcal{N}_y$ are consistent. We then compute their canonical interval solutions, and therefore construct a frame $T$ and a rectangle $m_i$ for each $i$. As in the last section, we continue to compute the digital region $b_i$ and try to find the connected component $c_i$ such that $\mathcal{M}(c_i) = m_i$. If such $c_i$ does not exist for some $i$, then $\mathcal{N}$ is inconsistent. Otherwise, set $\mathfrak{c} = \{c_i\}_{i=1}^n$. If $\mathfrak{c}$ is a solution, then $\mathcal{N}$ is consistent. Otherwise, as implied by Thm. 4, $\mathcal{N}$ must be inconsistent.

Fig. 13 gives the flowchart of the algorithm.

### 5.1  An $O(n^4)$ Consistency Checking Algorithm

In this section, we give a detailed description of our consistency checking algorithm.

**Step 1. Projective IA Networks**

For any basic CDC constraints $\{x_1 \delta_{12} x_2, x_2 \delta_{21} x_1\}$, the $x$- and $y$-projective IA relations $\rho_{12}^x$ and $\rho_{12}^y$ (Eq.s 11 and 12) can be computed in constant time. So the projective IA networks $\mathcal{N}_x$ and $\mathcal{N}_y$ can be constructed in $O(n^2)$ time. If $\mathcal{N}_x$ or $\mathcal{N}_y$ is inconsistent, which can be checked in cubic time, then $\mathcal{N}$ is inconsistent.

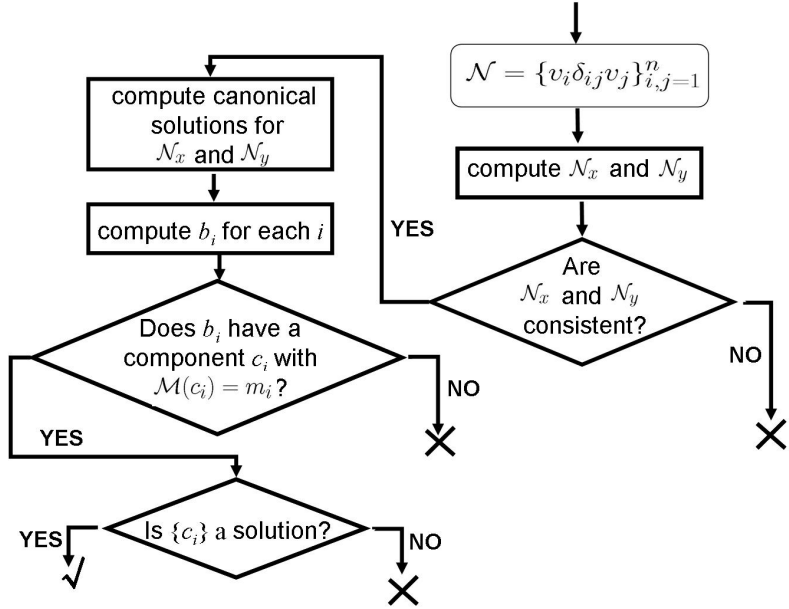

Figure 13: Flowchart of the main algorithm

**Step 2. Canonical Interval Solutions**

Suppose $\mathcal{N}_x$ and $\mathcal{N}_y$ are consistent. Their canonical solutions $\{I_i\}_{i=1}^n$ and $\{J_i\}_{i=1}^n$ can be constructed in cubic time.

Write $I_i = [x_i^-, x_i^+]$ and $J_i = [y_i^-, y_i^+]$. By definition of canonical interval solution, we know $x_i^-, x_i^+, y_i^-, y_i^+$ are integers between 0 and $2n-1$. Write $n_x = \max\{x_i^+\}$ and $n_y = \max\{y_i^+\}$. Set $T = [0, n_x] \times [0, n_y]$ as the frame and define $m_i = [x_i^-, x_i^+] \times [y_i^-, y_i^+]$ for each $i$.

**Step 3. Excluding Impossible Pixels**

As before, we write $D_i$ for the set of pixels in $m_i$ that are disallowed for $v_i$, i.e.

$$D_i = \{p_{st} \subseteq m_i : \text{there exist } j \neq i, \ \phi \text{ such that } p_{st} \subseteq m_j^\phi \text{ and } d_{ij}^\phi = 0\}$$

Equivalently, $p_{st} \in D_i$ if and only if $p_{st} \subseteq m_i \cap \bigcup \{m_j^\phi : d_{ij}^\phi = 0, j \neq i\}$. Set

$$b_i = \bigcup \{p_{st} \subseteq m_i : p_{st} \notin D_i\}$$

Clearly, $b_i^\circ \cap m_j^\phi$ is empty for any $j, \phi$ with $d_{ij}^\phi = 0$. Note that each $b_i$ can be computed in $O(n^2)$ time from $D_i$. To compute $D_i$, an intuitive method is checking for each pixel $p$ in $m_i$, and each tile $m_j^\phi$ of any reference object $m_j$, whether $p$ is in $m_j^\phi$. Note that there are at most $O(n^2)$ pixels contained in $m_i$ and at most $O(n)$ different $m_j^\phi$. This requires cubic time for a fixed $i$, and hence $O(n^4)$ time in total. Later, we will show this can be improved to $O(n^3)$.

### Step 4. Connected Components

We further compute connected components of $b_i$ for each $i$. Note that $b_i$ has at most one component whose mbr is $m_i$ (Lemma 16). If no such component exists for some $i$, then $\mathcal{N}$ is inconsistent. Otherwise, set $c_i$ for the connected component of $b_i$ such that $\mathcal{M}(c_i) = m_i$.

Applying a general Breadth-First Search algorithm, we can find all connected components of $b_i$ and determine if their mbrs are $m_i$ in $O(n^2)$ time. So, this step need only $O(n^3)$ time.

### Step 5. Checking A Possible Solution

The last step is then checking if $\{c_i\}_{i=1}^n$ is a solution of $\mathcal{N}$. Note that if the answer is yes, then $\{c_i\}_{i=1}^n$ is the maximal canonical solution of $\mathcal{N}$.

For each pair $c_i$ and $c_j$, we should check if $\text{dir}(c_i, c_j) = \delta_{ij}$. In other words, we should check for each $1 \leq \phi \leq 9$ whether the following equation holds

$$c_i^\circ \cap m_j^\phi = \varnothing \Leftrightarrow d_{ij}^\phi = 0 \tag{28}$$

When $d_{ij}^\phi = 0$, by the definition of $b_i$, we have $b_i^\circ \cap m_j^\phi = \varnothing$. Since $c_i$ is contained in $b_i$, we know $c_i^\circ \cap m_j^\phi = \varnothing$. The condition is always satisfied.

So to determine if $\text{dir}(c_i, c_j) = \delta_{ij}$, we need only check whether $c_i^\circ \cap m_j^\phi$ is nonempty for each $\phi$ with $d_{ij}^\phi = 1$.

Recall $c_i$ is a connected component of $b_i$ and $\mathcal{M}(c_i) = m_i$. If $m_i^\circ \cap m_j^\phi = \varnothing$, then $c_i^\circ \cap m_j^\phi = \varnothing$. The condition is violated and hence $\mathcal{N}$ is inconsistent. Note that each $m_i$ has been computed and each $m_j^\phi$ can be computed in constant time. Therefore, whether $m_i^\circ \cap m_j^\phi$ is empty can be checked in constant time for any $i, j, \phi$.

Suppose $m_i^\circ \cap m_j^\phi \neq \varnothing$. Write $m_i \cap m_j^\phi = [x^-, x^+] \times [y^-, y^+]$. To check if $c_i^\circ \cap m_j^\phi$ is nonempty, we need show there is a pixel $p$ which is contained in both $m_i \cap m_j^\phi$ and $c_i$. We need not check this for all pixels in $m_i \cap m_j^\phi$. Instead, we need only check whether $p_{kl} \subseteq c_i$ for all *boundary* pixels $p_{kl}$ of $m_i \cap m_j^\phi$ with $(k, l) \in H_1 \cup H_2$, where

$$\begin{aligned} H_1 &= \{(k,l) : k \in \{x^-, x^+ - 1\} \text{ and } y^- \leq l < y^+\}, \\ H_2 &= \{(k,l) : x^- \leq k < x^+ \text{ and } l \in \{y^-, y^+ - 1\}\}. \end{aligned}$$

We justify the above statement as follows. If $m_i \cap m_j^\phi = m_i$, then by $\mathcal{M}(c_i) = m_i$ we know there exists a boundary pixel which is contained in $c_i$. Otherwise, $m_i \cap m_j^\phi$ is a rectangle strictly contained in $m_i$, which means $c_i$ contains a pixel $p$ out of in $m_i \cap m_j^\phi$. If $c_i$ contains no boundary pixel of $m_i \cap m_j^\phi$, then, because it is connected, $c_i$ contains no pixel of $m_i \cap m_j^\phi$ at all.

Since $H_1 \cup H_2$ contains $O(n)$ pixels and checking if a pixel is contained in $c_i$ needs constant time, $\mathsf{dir}(c_i, c_j) = \delta_{ij}$ can be checked in $O(n)$ time.

In conclusion, we can determine in cubic time whether $\{c_i\}_{i=1}^n$ is a solution of $\mathcal{N}$. Now, since only Step 3 needs at most $O(n^4)$ time, the algorithm determines the consistency of a basic CDC network in $O(n^4)$ time.

## 5.2 A Cubic Improvement

In this subsection, we improve the main algorithm to cubic. This is achieved by an $O(n^2)$ improvement for computing each $D_i$ (see Eq. 24) in Step 3. As a consequence, all $b_i$ can be computed in cubic time.

Suppose $T = [0, n_x] \times [0, n_y]$. Then any digital region $a \subseteq T$ can be represented as an $n_x \times n_y$ Boolean matrix $B(a)$ as follows:

$$B(a)[k, l] = \left\{ \begin{array}{ll} 1, & \text{if } p_{kl} \in a; \\ 0, & \text{otherwise.} \end{array} \right. \qquad (29)$$

**Example 3.** *The left of Fig. 14 illustrates a digital region $a$ contained in $T = [0, 4] \times [0, 4]$, and the right of Fig. 14 shows the Boolean matrix $B(a)$ that represents $a$.*

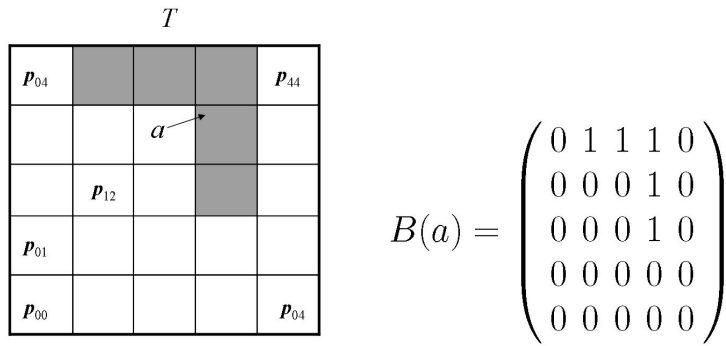

Figure 14: A digital region $a$ and the Boolean matrix $B(a)$

It is worthy noting that we adopt a different way to address the elements of $B(a)$. For example, the lower left corner of $B(a)$ in Fig. 14 is addressed as the (0,0)-element, i.e. $B(a)[0, 0]$, instead of the (5,1)-element.

For each $i$, recall $D_i$ is the set of pixels in $m_i$ that are disallowed for $v_i$ (cf. Eq. 24 or Step 3 of the main algorithm). For simplicity, we write $P_i$ for the matrix that represents the digital region which consists of pixels in $D_i$. Similarly, we write $B_j$ for the matrix that represents the rectangle $m_j$, and write $B_j^\phi$ for the matrix that represents $T \cap m_j^\phi$. In case $T \cap m_j^\phi$ contains no pixel, we assume it is represented by the zero matrix.

Boolean operations $\wedge$ and $\vee$ can be defined on Boolean matrices in a natural way. By definition of $D_i$ (see Eq. 24), we have

$$P_i = B_i \wedge \bigvee \{B_j^\phi : d_{ij}^\phi = 0 \text{ and } j \neq i\}. \tag{30}$$

Adding up all $B_j^\phi$ with $d_{ij}^\phi = 0$ and $j \neq i$, we obtain an integer matrix $Q_i$, i.e.

$$Q_i = \sum \{B_j^\phi : d_{ij}^\phi = 0 \text{ and } j \neq i\}. \tag{31}$$

Easy to see, the following equation holds:

$$P_i[k,l] = \begin{cases} 1, & \text{if } Q_i[k,l] > 0 \text{ and } B_i[k,l] = 1; \\ 0, & \text{otherwise.} \end{cases} \tag{32}$$

Note that $T \cap m_j^\phi$ is a (may be degenerative) rectangle. We next show that this property plays an important role in the improvement. To this end, we introduce the following operations on matrices.

**Definition 14** (cumulative matrix and difference matrix). For a matrix $N$, we define its *cumulative matrix* as

$$\mathsf{acc}(N)[k,l] = \sum_{t=0}^{k} N[t,l]. \tag{33}$$

and its *difference matrix* as

$$\mathsf{diff}(N)[k,l] = \begin{cases} N[k,l], & \text{if } k = 0; \\ N[k,l] - N[k-1,l], & \text{otherwise.} \end{cases} \tag{34}$$

The cumulative matrix can be computed column by column. We first add the first column to the second one, and then add the updated second column to the third, *etc.* In this way, cumulative matrix $\mathsf{acc}(N)$ can be computed linearly in the number of elements of $N$.

It is easy to verify that $N = \mathsf{acc}(\mathsf{diff}(N))$. Since the difference operation is additive, *i.e.* $\mathsf{diff}(N_1 + N_2) = \mathsf{diff}(N_1) + \mathsf{diff}(N_2)$, we have

$$\begin{aligned} Q_i &= \mathsf{acc}(\mathsf{diff}(Q_i)) \\ &= \mathsf{acc}(\mathsf{diff}(\sum \{B_j^\phi : d_{ij}^\phi = 0 \text{ and } j \neq i\})) \\ &= \mathsf{acc}(\sum \{\mathsf{diff}(B_j^\phi) : d_{ij}^\phi = 0 \text{ and } j \neq i\}) \end{aligned}$$

We next show each $Q_i$ can be computed in quadratic time. To this end, we need the following lemma, which asserts that the number of non-zero elements in $\mathsf{diff}(B_j^\phi)$ is of order $O(n)$.

**Lemma 19.** *The number of non-zero elements in* $\mathsf{diff}(B_j^\phi)$ *is less than* $4n$.

*Proof.* If $B_j^\phi$ is the zero matrix, then $\mathsf{diff}(B_j^\phi)$ is also the zero matrix. Otherwise, the non-zero elements in $B_j^\phi$ compose a rectangle, *i.e.*, there exist $0 \leq x_1 \leq x_2 < n_x$ and $0 \leq y_1 \leq y_2 < n_y$ s.t.

$$B_j^\phi[k,l] = \begin{cases} 1, & \text{if } x_1 \leq k \leq x_2 \text{ and } y_1 \leq l \leq y_2; \\ 0, & \text{otherwise.} \end{cases}$$

Easy to prove,

$$\mathsf{diff}(B_j^\phi)[k,l] = \begin{cases} 1, & \text{if } k = x_1 \quad \text{and } y_1 \leq l \leq y_2; \\ -1, & \text{if } k = x_2 + 1 \quad \text{and } y_1 \leq l \leq y_2; \\ 0, & \text{otherwise.} \end{cases}$$

So if $x_2 < n_x - 1$, there are $(y_2 - y_1 + 1)$ '1's and '-1's in $\mathsf{diff}(B_j^\phi)$; otherwise $x_2 = n_x - 1$, there are $(y_2 - y_1 + 1)$ '1's and none '-1's, with other elements being zeros. So there are at most $2 \times (y_2 - y_1 + 1) \leq 2 \times n_y \leq 4n$ non-zero elements in $\mathsf{diff}(B_j^\phi)$. □

Since $1 \leq j \leq n$ and $1 \leq \phi \leq 9$, there are at most $9n$ $B_j^\phi$ need to consider. By Lemma 19, we know $\sum\{\mathsf{diff}(B_j^\phi) : d_{ij}^\phi = 0\}$ can be computed in quadratic time. As it is a matrix with $n_x \times n_y$ elements, computing its cumulative matrix, *viz.* $Q_i$, also needs quadratic time.

By Eq. 32, we know $P_i$, the matrix representation of $D_i$, can be computed in quadratic time. As a consequence, each $b_i$ can be computed in $O(n^2)$, instead of $O(n^3)$, time. In this way, we improve Step 3 of the main algorithm from $O(n^4)$ to cubic time.

**Example 1 (continued)**
In our running example, for $m_1$, there are four different $m_j^\phi$ such that $B_j^\phi$ is not the zero matrix and $d_{1j}^\phi = 0$, *viz.* $m_2^5 = m_2$, $m_2^8$, $m_2^9$, $m_3^4$. We have

$$B_2^5 = \begin{pmatrix} 0 & 0 & 0 & 0 & 0 \\ 0 & 0 & 0 & 0 & 0 \\ 1 & 1 & 0 & 0 & 0 \\ 1 & 1 & 0 & 0 & 0 \\ 0 & 0 & 0 & 0 & 0 \end{pmatrix} \quad \mathsf{diff}(B_2^5) = \begin{pmatrix} 0 & 0 & 0 & 0 & 0 \\ 0 & 0 & 0 & 0 & 0 \\ 1 & 0 & -1 & 0 & 0 \\ 1 & 0 & -1 & 0 & 0 \\ 0 & 0 & 0 & 0 & 0 \end{pmatrix}$$

$$B_2^8 = \begin{pmatrix} 0 & 0 & 0 & 0 & 0 \\ 0 & 0 & 0 & 0 & 0 \\ 0 & 0 & 0 & 0 & 0 \\ 0 & 0 & 0 & 0 & 0 \\ 1 & 1 & 0 & 0 & 0 \end{pmatrix} \quad \mathsf{diff}(B_3^4) = \begin{pmatrix} 0 & 0 & 0 & 0 & 0 \\ 0 & 0 & 0 & 0 & 0 \\ 0 & 0 & 0 & 0 & 0 \\ 0 & 0 & 0 & 0 & 0 \\ 1 & 0 & -1 & 0 & 0 \end{pmatrix}$$

$$B_2^9 = \begin{pmatrix} 0 & 0 & 0 & 0 & 0 \\ 0 & 0 & 0 & 0 & 0 \\ 0 & 0 & 0 & 0 & 0 \\ 0 & 0 & 0 & 0 & 0 \\ 0 & 0 & 1 & 1 & 1 \end{pmatrix} \quad \mathsf{diff}(B_2^9) = \begin{pmatrix} 0 & 0 & 0 & 0 & 0 \\ 0 & 0 & 0 & 0 & 0 \\ 0 & 0 & 0 & 0 & 0 \\ 0 & 0 & 0 & 0 & 0 \\ 0 & 0 & 1 & 0 & 0 \end{pmatrix}$$

$$B_3^4 = \begin{pmatrix} 0 & 0 & 0 & 0 & 0 \\ 1 & 1 & 1 & 0 & 0 \\ 1 & 1 & 1 & 0 & 0 \\ 1 & 1 & 1 & 0 & 0 \\ 1 & 1 & 1 & 0 & 0 \end{pmatrix} \quad \mathsf{diff}(B_3^4) = \begin{pmatrix} 0 & 0 & 0 & 0 & 0 \\ 1 & 0 & 0 & -1 & 0 \\ 1 & 0 & 0 & -1 & 0 \\ 1 & 0 & 0 & -1 & 0 \\ 1 & 0 & 0 & -1 & 0 \end{pmatrix}$$

Summing up all corresponding $\mathsf{diff}(B_j^\phi)$ and calculating its cumulative matrix, we obtain $Q_1$.

$$Q_1 = \mathsf{acc}\left(\begin{pmatrix} 0 & 0 & 0 & 0 & 0 \\ 1 & 0 & 0 & -1 & 0 \\ 1 & 0 & 0 & -1 & 0 \\ 1 & 0 & 0 & -1 & 0 \\ 2 & 0 & 0 & -1 & 0 \end{pmatrix}\right) = \begin{pmatrix} 0 & 0 & 0 & 0 & 0 \\ 1 & 1 & 1 & 0 & 0 \\ 1 & 1 & 1 & 0 & 0 \\ 1 & 1 & 1 & 0 & 0 \\ 2 & 2 & 2 & 1 & 1 \end{pmatrix}$$

By Eq. 32, we know $P_1[k,l] = \min\{B_1[k,l], Q_1[k,l]\}$. Therefore,

$$P_1 = \min\left\{\begin{pmatrix} 0 & 1 & 1 & 1 & 0 \\ 0 & 1 & 1 & 1 & 0 \\ 0 & 1 & 1 & 1 & 0 \\ 0 & 0 & 0 & 0 & 0 \\ 0 & 0 & 0 & 0 & 0 \end{pmatrix}, \begin{pmatrix} 0 & 0 & 0 & 0 & 0 \\ 1 & 1 & 1 & 0 & 0 \\ 1 & 1 & 1 & 0 & 0 \\ 1 & 1 & 1 & 0 & 0 \\ 2 & 2 & 2 & 1 & 1 \end{pmatrix}\right\} = \begin{pmatrix} 0 & 0 & 0 & 0 & 0 \\ 0 & 1 & 1 & 0 & 0 \\ 0 & 1 & 1 & 0 & 0 \\ 0 & 0 & 0 & 0 & 0 \\ 0 & 0 & 0 & 0 & 0 \end{pmatrix}.$$

Finally, we get $b_1$ by Eq. 25, shown in Fig. 12, which is clearly connected. This shows $c_1 = b_1$.

### 5.3 Beyond Basic CDC Constraints

In the above two subsections, we have shown that the consistency of a basic network of CDC constraints can be determined in cubic time. Using backtracking method, we immediately know that the consistency satisfaction problem of CDC is an NP problem.

**Lemma 20.** *The consistency satisfaction problem of CDC is an NP problem.*

*Proof.* Let $\mathcal{C} = \{v_i c_{ij} v_j\}_{i,j=1}^n$ be a set of CDC constraints. To determine if $\mathcal{C}$ is consistent, we need only branch each non-basic constraint $c_{ij}$, and then call our cubic algorithm to solve the basic network of CDC constraints. □

This problem is also NP-hard.

**Lemma 21.** *The consistency satisfaction problem of CDC is NP-hard.*

*Proof.* We prove this by reducing a known NP-hard problem to the consistency satisfaction problem of CDC. Let $\mathcal{A}$ be the JEPD set of IA relations

$$\{\mathsf{p} \cup \mathsf{m} \cup \mathsf{pi} \cup \mathsf{mi}, \mathsf{o} \cup \mathsf{s} \cup \mathsf{d} \cup \mathsf{f} \cup \mathsf{eq} \cup \mathsf{fi} \cup \mathsf{si} \cup \mathsf{di} \cup \mathsf{oi}\}.$$

It can be proved that reasoning with $\mathcal{A}$ is already NP-hard. Actually, the NP-hardness of $\mathcal{A}$ is guaranteed by the work of Krokhin, Jeavons, and Jonsson [15]. We need only to show $\mathcal{A}$ is not contained in any of the eighteen maximal tractable subclasses of IA (see [15, Table III]). The verification is straightforward and omitted here.

Reasoning with IA relations in $\mathcal{A}$ can be easily reduced to reasoning with CDC. For a set of IA constraints

$$\mathcal{N} = \{v_i \lambda_{ij} v_j\}_{i,j=1}^n, \ (\lambda_{ij} \in \mathcal{A}),$$

it is easy to see that $\mathcal{N}$ is satisfiable if and only if the set of CDC constraints

$$\mathcal{N}^* = \{v_i \varphi_{ij} v_j\}_{i,j=1}^n$$

is consistent, where $\varphi_{ij}$ is the disjunction of the basic CDC relations in

$$\left\{ \begin{pmatrix} 0 & 0 & 0 \\ 1 & 0 & 0 \\ 0 & 0 & 0 \end{pmatrix}, \begin{pmatrix} 0 & 0 & 0 \\ 0 & 0 & 1 \\ 0 & 0 & 0 \end{pmatrix} \right\}$$

if $\lambda_{ij} = \mathsf{p} \cup \mathsf{m} \cup \mathsf{pi} \cup \mathsf{mi}$, and is the disjunction of the basic CDC relations in

$$\left\{ \begin{pmatrix} 0 & 0 & 0 \\ 0 & 1 & 0 \\ 0 & 0 & 0 \end{pmatrix}, \begin{pmatrix} 0 & 0 & 0 \\ 1 & 1 & 0 \\ 0 & 0 & 0 \end{pmatrix}, \begin{pmatrix} 0 & 0 & 0 \\ 0 & 1 & 1 \\ 0 & 0 & 0 \end{pmatrix}, \begin{pmatrix} 0 & 0 & 0 \\ 1 & 1 & 1 \\ 0 & 0 & 0 \end{pmatrix} \right\},$$

otherwise.

This shows that reasoning with CDC is at least as hard as reasoning with $\mathcal{A}$. Therefore, reasoning with CDC is also NP-hard. □

As a corollary, we know

**Theorem 5.** *Reasoning with CDC is an NP-Complete problem.*

# 6 The Pairwise Consistency Problem and The Weak Composition Problem

Our main algorithm can be applied to solve the two special subproblems of reasoning with CDC, *i.e.* the pairwise consistency problem and the weak composition problem. These two subproblems have been considered in [3] and [32], respectively. In this section, we will compare their results with ours.

## 6.1 The Pairwise Consistency Problem

Given that you know the relation of $a$ to $b$, what about that of $b$ to $a$? Mathematically speaking, this is the converse problem. The relation of $b$ to $a$ is the converse of that of $a$ to $b$, and vice versa. Suppose $\mathcal{B}$ is a set of JEPD relations on $D$. The qualitative calculus $\langle \mathcal{B} \rangle$ is not necessarily closed under converse. This means, $\alpha^\sim$ could be a relation outside of $\langle \mathcal{B} \rangle$ despite $\alpha \in \langle \mathcal{B} \rangle$.

As for CDC, we have shown in Fig. 3 that a basic CDC relation may have more than one 'converses,' where for two basic relations $\alpha, \beta$, we say $\alpha$ is a *converse* of $\beta$ (in CDC) if $\{v_1 \alpha v_2, v_2 \beta v_1\}$ is consistent.

The pairwise consistency problem in CDC is the problem of deciding if $\{v_1 \delta_{12} v_2, v_2 \delta_{21} v_1\}$ is satisfiable for a pair of basic CDC relations $\delta_{12}$ and $\delta_{21}$. This problem has been discussed by Cicerone and di Felice [3], where they identified 2004 consistent pairs of basic CDC relations.

We next apply our main algorithm to solve the pairwise consistency problem. The first step computes $\rho_{12}^x = \iota^x(\delta_{12}) \cap \iota^x(\delta_{21})^\sim$ and $\rho_{12}^y = \iota^y(\delta_{12}) \cap \iota^y(\delta_{21})^\sim$. If $\rho_{12}^x$ or $\rho_{12}^y$ is empty, then the program stops and returns 'inconsistent.' Otherwise, we go to Step 2. We construct the canonical solutions $\{I_1, I_2\}$ and $\{J_1, J_2\}$ to $\rho_{12}^x$ and $\rho_{12}^y$, respectively. Set $m_i = I_i \times J_i$ for $i = 1, 2$. Write $I_i = [x_i^-, x_i^+]$ and $J_i = [y_i^-, y_i^+]$. Denote $n_x = \max\{x_1^+, x_2^+\}$, $n_y = \max\{y_1^+, y_2^+\}$. Clearly, $1 \leq n_x, n_y \leq 3$. Set $T = [0, n_x] \times [0, n_y]$. Then $T \subseteq [0, 3] \times [0, 3]$. Continuing as described in the main algorithm, we will determine if $\{v_1 \delta_{12} v_2, v_2 \delta_{21} v_1\}$ is consistent, and find the maximal canonical solution in case it is consistent.

A specialized algorithm is implemented. We obtain in total 757 consistent pairs of basic CDC relations. Among these consistent pairs $(\delta, \delta')$, one $\delta$ may correspond to multiple $\delta'$, *i.e.* $\delta$ may have multiple converses. In fact, among the 218 basic CDC relations, 119 have unique converse, 68 have two converses, 6 have four converses, 20 have eight converses, 4 have thirty converses, and one (*viz.* the one such that $d_{ij} = 1$ iff $i = j = 2$) has 198 converses.

Our result is unexpectedly different from that of [3], where Cicerone and di Felice obtained 2004 consistent pairs of basic CDC relations. A careful examination, however, shows that a similar but different model was used in [3].

When defining the direction relation matrix $\text{dir}(a, b) = [d_{ij}]_{i,j=1}^3$ of $a$ to $b$ (see Dfn. 4), we require $d_{ij} = 1$ if and only if $a^\circ \cap b_{ij} \neq \varnothing$. While in [3], the $(ij)$-entry $d_{ij}$ is 1 if and only if $a$ has nonempty intersection with $b_{ij}$, *i.e.* $d_{ij} = 1$ iff $a \cap b_{ij} \neq \varnothing$. We call these two definitions the *interior-based* and, respectively, the *closure-based* direction relation matrix.

In the following, we argue that the interior-based definition is more coherent than the closure-based one.

First, though the original definition of Direction Relation Matrix (DRM) [11] does not mention that $d_{ij} = 1$ iff $a$ has a common interior point with the tile $b_{ij}$, Goyal and Egenhofer defined in [12] the detailed direction relation matrix of $a$ to $b$ as a numerical matrix $\text{dir}^*(a, b) = [d_{ij}^*]_{i,j=1}^3$, where $d_{ij}^*$ is interpreted as the ratio of the area of $a \cap b_{ij}$ and $a$. This is a natural extension of the interior-based direction relation matrix. From $\text{dir}^*(a, b) = [d_{ij}^*]_{i,j=1}^3$, we can obtain the coarse direction relation matrix $\text{dir}(a, b) = [d_{ij}]_{i,j=1}^3$ by setting $d_{ij} = 1$ iff $d_{ij}^* > 0$.

Second, the qualitative calculus introduced by the interior-based direction relation matrix is more desirable. For example, it is a natural requirement that the identity relation is contained in a unique basic CDC relation. For the interior-based definition, we have

$$\mathsf{dir}(a,a) = \begin{pmatrix} 0 & 0 & 0 \\ 0 & 1 & 0 \\ 0 & 0 & 0 \end{pmatrix}.$$

for any connected region $a$. The following example, however, shows that this is not the case for the closure-based direction relation matrix.

**Example 4.** *Take $a$ as the square $[-1,1] \times [-1,1]$, and take $a'$ as the unit disk centered at $(0,0)$. Then $\mathcal{M}(a') = \mathcal{M}(a) = a$, but $\mathsf{dir}(a,a) \neq \mathsf{dir}(a',a')$ if we take the closure-based definition.*

$$\mathsf{dir}(a,a) = \begin{pmatrix} 1 & 1 & 1 \\ 1 & 1 & 1 \\ 1 & 1 & 1 \end{pmatrix}, \quad \mathsf{dir}(a',a') = \begin{pmatrix} 0 & 1 & 0 \\ 1 & 1 & 1 \\ 0 & 1 & 0 \end{pmatrix}$$

Moreover, the interior-based definition is also consistent with the one adopted in [32, 33].

## 6.2 The Weak Composition Problem

The notion of weak composition plays a very important role in qualitative spatial and temporal reasoning [1, 5, 18, 28]. For two basic CDC relations $\alpha, \beta$, the weak composition $\alpha \circ_w \beta$ of $\alpha$ and $\beta$ is defined to be the smallest relation in the CDC algebra which contains the composition $\alpha \circ \beta$. Since CDC is a Boolean algebra, $\alpha \circ_w \beta$ is the union of all basic CDC relations it contains. For a basic CDC relation $\gamma$, it is easy to prove that

$$\gamma \subseteq \alpha \circ_w \beta \Leftrightarrow \gamma \cap (\alpha \circ \beta) \neq \varnothing. \tag{35}$$

Note that $\gamma \cap (\alpha \circ \beta)$ is nonempty iff the following set of basic CDC constraints

$$\mathcal{C} = \{v_1 \alpha v_2, v_2 \beta v_3, v_1 \gamma v_3\} \tag{36}$$

is consistent. We note that $\mathcal{C}$ is not a complete network. The constraint of, say, $v_2$ to $v_1$, is not specified. According to the previous subsection, $\alpha$ may have multiple converses. To apply our main algorithm, we need to extend $\mathcal{C}$ to a complete network:

$$\mathcal{C}^* = \{v_1 \alpha v_2, v_2 \alpha' v_1, v_2 \beta v_3, v_3 \beta' v_2, v_1 \gamma v_3, v_3 \gamma' v_1\} \tag{37}$$

We then call our main algorithm to determine if the above completed network is consistent. If the answer is 'yes' for some $\alpha', \beta', \gamma'$, then $\gamma$ is contained in the weak composition of $\alpha$ and $\beta$. Note that we need only to apply the main algorithm to those $\alpha', \beta', \gamma'$ such that $(\alpha, \alpha'), (\beta, \beta'), (\gamma, \gamma')$ are consistent pairs.

The above algorithm was implemented, and its codes are available from the authors.

The same problem has been considered in [32], where Skiadopoulos and Koubarakis gave an algorithm to compute the weak composition. The main idea is to compute the weak composition progressively.

**Definition 15** (single-tile, multi-tile, component). A basic CDC relation is called *single-tile* if its matrix has only one nonzero entry, and called *multi-tile* otherwise. We say a single-tile relation $[s_{ij}]_{i,j=1}^3$ is a *component* of a multi-tile relation $[d_{ij}]_{i,j=1}^3$, if $s_{ij} \leq d_{ij}$ for $1 \leq i,j \leq 3$.

The weak composition of two single-tile relations has been computed by Goyal [10]. Upon this, Theorem 1 of [32] establishes a rule for computing the weak composition of a single-tile relation and a basic relation. The correctness of this theorem is confirmed by our algorithm. Furthermore, Theorem 2 of [32] then gives a rule to compute the weak composition of two multi-tile relations.

**Definition 16** (decomposability). Let $\alpha, \beta$ are two basic CDC relations. Suppose $\alpha_1, \alpha_2, \cdots, \alpha_k$ are the component single-tile relations of $\alpha$. We say a basic relation $\gamma$ is *decomposable with respect to* $(\alpha, \beta)$, if $\gamma$ is the join of $k$ direction relation matrices $\gamma_s$, where each $\gamma_s$ is a basic relation contained in $\alpha_s \circ_w \beta$.

The following rule was used in [32] to compute the weak composition between $\alpha$ and any other basic CDC relation $\beta$.

**[32, Theorem 2]** *A basic relation $\gamma$ is contained in $\alpha \circ_w \beta$ if and only if $\gamma$ is decomposable with respect to $(\alpha, \beta)$.*

We next give two examples to show that this rule is incorrect in some cases.

**Example 5.** *For basic relations $\alpha, \beta, \gamma$ in Eq. 38,*

$$\alpha = \begin{pmatrix} 0 & 0 & 0 \\ 1 & 0 & 0 \\ 1 & 0 & 0 \end{pmatrix} \quad \beta = \begin{pmatrix} 0 & 0 & 1 \\ 0 & 0 & 1 \\ 0 & 0 & 0 \end{pmatrix} \quad \gamma = \begin{pmatrix} 1 & 0 & 1 \\ 1 & 0 & 1 \\ 1 & 1 & 1 \end{pmatrix} \tag{38}$$

*we assert that $\gamma$ is not decomposable with respect to $(\alpha, \beta)$ (see Lemma 23 below). According to [32, Theorem 2], this will imply that $\gamma \not\subseteq \alpha \circ_w \beta$, i.e. $\{v_1 \alpha v_2, v_2 \beta v_3, v_1 \gamma v_3\}$ is inconsistent. The consistency of the above set is, however, confirmed by our algorithm. Fig. 15 (left) gives a configuration of three regions $\{a, b, c\}$ which satisfies the constraints.*

We next show $\gamma$ is not decomposable w.r.t. $(\alpha, \beta)$. First of all, it is easy to see that $\alpha_1$ and $\alpha_2$ in Eq. 39 are the only component single-tile relations of $\alpha$ in Eq. 38.

$$\alpha_1 = \begin{pmatrix} 0 & 0 & 0 \\ 0 & 0 & 0 \\ 1 & 0 & 0 \end{pmatrix} \quad \alpha_2 = \begin{pmatrix} 0 & 0 & 0 \\ 1 & 0 & 0 \\ 0 & 0 & 0 \end{pmatrix} \tag{39}$$

The next lemma characterizes when a basic relation is contained in the weak composition of $\alpha_s$ and $\beta$ for $s = 1, 2$.

**Lemma 22.** *A basic relation $\pi$ is contained in $\alpha_s \circ_w \beta$ if and only if $\pi$ is a valid matrix whose component single-tile relations are also component single-tile relations of $\pi_s$, where $\pi_s$ ($s = 1, 2$) are matrices in Eq. 40.*

$$\pi_1 = \begin{pmatrix} 0 & 0 & 0 \\ 1 & 1 & 1 \\ 1 & 1 & 1 \end{pmatrix} \quad \pi_2 = \begin{pmatrix} 1 & 1 & 1 \\ 1 & 1 & 1 \\ 0 & 0 & 0 \end{pmatrix} \tag{40}$$

*Proof.* The proof is straightforward. We illustrate the weak composition in Fig. 15 (middle and right). Take $\alpha_1$ in Eq. 39 as example. In the middle of Fig. 15, there are six pixels marked "$a$." For each $a$, we have $a\alpha_1 b$ and $b\beta c$. Moreover, $\mathsf{dir}(a, c)$ is a single-tile relation for each of the six $a$. These single-tile relations are components of $\pi_1$. The case of $\alpha_2$ is similar and illustrated in the right of Fig. 15. □

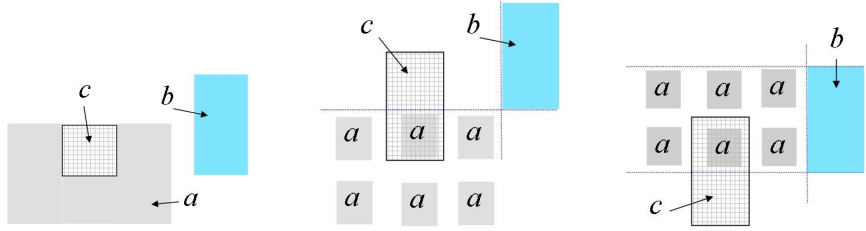

Figure 15: Illustrations of evidences of weak compositions $\alpha \circ_w \beta$ (left), $\alpha_1 \circ_w \beta$ (middle), $\alpha_2 \circ_w \beta$ (right)

Then we arrive at our conclusion.

**Lemma 23.** *$\gamma$ is not decomposable with respect to $(\alpha, \beta)$.*

*Proof.* Suppose $\gamma = \gamma_1 \vee \gamma_2$, and $\gamma_s \subseteq \alpha_s \circ_w \beta$ ($s = 1, 2$). This means, by Lemma 22, $\gamma_s$ is a valid matrix whose component single-tile relations are all $\pi_s$'s. For $1 \le i, j \le 3$, we have $\gamma[i, j] = \gamma_1[i, j] \vee \gamma_2[i, j]$, and $\gamma_s[i, j] \le \pi_s[i, j]$ for $s = 1, 2$. Clearly, $\gamma_1[1, 1] = \gamma_1[1, 3] = 0$ and hence $\gamma_2[1, 1] = \gamma_2[1, 3] = 1$. Similarly, $\gamma_2[1, 2] = \gamma_2[2, 2] = \gamma_2[3, 2] = 0$. Therefore, $\gamma_2$ is not 4-connected, hence not a direction relation matrix. This shows that $\gamma$ is not decomposable. □

The following example provides a set of three basic CDC constraints which is inconsistent but satisfies the condition of [32, Theorem 2].

**Example 6.** *For basic relations $\alpha', \beta', \gamma'$ as in Eq. 41,*

$$\alpha' = \begin{pmatrix} 1 & 1 & 0 \\ 1 & 0 & 0 \\ 1 & 1 & 0 \end{pmatrix} \quad \beta' = \begin{pmatrix} 0 & 0 & 0 \\ 1 & 0 & 0 \\ 1 & 1 & 0 \end{pmatrix} \quad \gamma' = \begin{pmatrix} 1 & 1 & 0 \\ 0 & 1 & 0 \\ 1 & 1 & 0 \end{pmatrix} \tag{41}$$

we assert that $\gamma'$ is decomposable with respect to $\alpha'$ and $\beta'$. Let $\alpha'_s$ ($s = 1, \cdots, 5$) (the middle matrices of Eqs. 42-46) be the component single-tiled relations of $\alpha'$. It is easy to see $\gamma'_s \subseteq \alpha'_s \circ_w \beta'$ (Eqs. 42-46). Evidences are given in Fig. 16, where a configuration $a\alpha'_s b, b\beta'c, a\gamma'_s c$ is given for each $s$. Note that $\gamma'$ is the join of $\gamma'_s$ for $s = 1, \cdots, 5$. According to Theorem 2 of [32], $\gamma'$ should be contained in $\alpha' \circ \beta'$, i.e.

$$\Gamma' = \{v_1 \alpha' v_2, v_2 \beta' v_3, v_1 \gamma' v_3\}$$

should be consistent. Our algorithm, however, shows that $\Gamma'$ is not consistent. An argument is also given in Lemma 24.

$$\gamma'_1 = \begin{pmatrix} 1 & 0 & 0 \\ 0 & 0 & 0 \\ 0 & 0 & 0 \end{pmatrix} \subseteq \begin{pmatrix} 1 & 0 & 0 \\ 0 & 0 & 0 \\ 0 & 0 & 0 \end{pmatrix} \circ_w \begin{pmatrix} 0 & 0 & 0 \\ 1 & 0 & 0 \\ 1 & 1 & 0 \end{pmatrix} \quad (42)$$

$$\gamma'_2 = \begin{pmatrix} 0 & 0 & 0 \\ 0 & 0 & 0 \\ 1 & 0 & 0 \end{pmatrix} \subseteq \begin{pmatrix} 0 & 0 & 0 \\ 1 & 0 & 0 \\ 0 & 0 & 0 \end{pmatrix} \circ_w \begin{pmatrix} 0 & 0 & 0 \\ 1 & 0 & 0 \\ 1 & 1 & 0 \end{pmatrix} \quad (43)$$

$$\gamma'_3 = \begin{pmatrix} 0 & 0 & 0 \\ 0 & 0 & 0 \\ 1 & 0 & 0 \end{pmatrix} \subseteq \begin{pmatrix} 0 & 0 & 0 \\ 0 & 0 & 0 \\ 1 & 0 & 0 \end{pmatrix} \circ_w \begin{pmatrix} 0 & 0 & 0 \\ 1 & 0 & 0 \\ 1 & 1 & 0 \end{pmatrix} \quad (44)$$

$$\gamma'_4 = \begin{pmatrix} 0 & 0 & 0 \\ 0 & 0 & 0 \\ 0 & 1 & 0 \end{pmatrix} \subseteq \begin{pmatrix} 0 & 0 & 0 \\ 0 & 0 & 0 \\ 0 & 1 & 0 \end{pmatrix} \circ_w \begin{pmatrix} 0 & 0 & 0 \\ 1 & 0 & 0 \\ 1 & 1 & 0 \end{pmatrix} \quad (45)$$

$$\gamma'_5 = \begin{pmatrix} 0 & 1 & 0 \\ 0 & 1 & 0 \\ 0 & 0 & 0 \end{pmatrix} \subseteq \begin{pmatrix} 0 & 1 & 0 \\ 0 & 0 & 0 \\ 0 & 0 & 0 \end{pmatrix} \circ_w \begin{pmatrix} 0 & 0 & 0 \\ 1 & 0 & 0 \\ 1 & 1 & 0 \end{pmatrix} \quad (46)$$

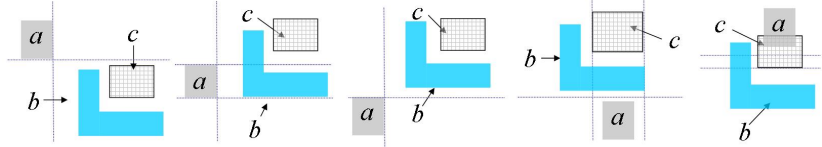

Figure 16: Illustrations of evidences of weak compositions $\alpha'_s \circ_w \beta'$ ($s = 1, \cdots, 5$)

**Lemma 24.** $\Gamma' = \{v_1 \alpha' v_2, v_2 \beta' v_3, v_1 \gamma' v_3\}$ is inconsistent.

*Proof.* Suppose $\mathfrak{a} = \{a, b, c\}$ is a solution to $\Gamma'$. We construct the frame $S(\mathfrak{a})$ and the cell set $C(\mathfrak{a})$ of $\mathfrak{a}$ (see Dfn. 9). Since $(b, c)$ is an instance of $\beta'$, we know $(\mathcal{M}(b), \mathcal{M}(c))$ is an instance of $\iota^x(\beta') \otimes \iota^y(\beta') = (\mathsf{o} \cup \mathsf{fi}) \otimes (\mathsf{o} \cup \mathsf{fi})$, *i.e.* $(\mathcal{M}(b), \mathcal{M}(c))$ is an instance of one of the four basic RA relations

$$\mathsf{o} \otimes \mathsf{o}, \mathsf{fi} \otimes \mathsf{o}, \mathsf{o} \otimes \mathsf{fi}, \mathsf{fi} \otimes \mathsf{fi}.$$

Take o ⊗ o as example. By $\alpha'[2,2] = \alpha'[1,3] = \alpha'[2,3] = \alpha'[3,3] = 0$, we know $a° \cap b_{ij} = \varnothing$ for $(ij) = (22),(13),(23),(33)$. Similarly, by $\gamma'[2,1] = \gamma'[1,3] = \gamma'[2,3] = \gamma'[3,3] = 0$, we know $a° \cap c_{ij} = \varnothing$ for $(ij) = (21),(13),(23),(33)$. Therefore, the impossible regions separate the plane into two disconnected pieces, one is above the impossible regions, the other below (see Fig. 17). Since $\gamma'[2,2] = 1$, i.e. $a° \cap \mathcal{M}(c) \neq \varnothing$, we know $a$ has an interior point $P$ which belongs to $\mathcal{M}(c)$, hence, the upper piece. By $\alpha'[3,2] = 1$, i.e. $a° \cap b_{32} \neq \varnothing$, we know $a$ has an interior point $Q$ which belongs to the lower piece. This is impossible since we assume $a$ is a connected region. □

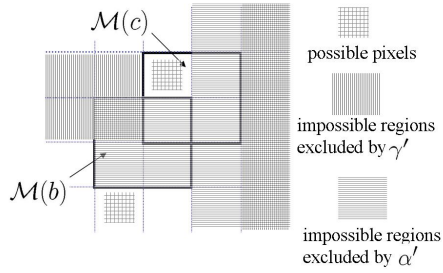

Figure 17: Illustration of Lemma 24

These two examples show that the weak composition algorithm proposed in [32] is not always correct.

# 7 Consistency Checking for Two Variants of CDC

The CDC algebra introduced in Dfn. 4 requires regions to be connected. This calculus has two variants in the literature. One, as introduced in [33], deals with cardinal direction relations between possibly disconnected regions; the other, as originally proposed in [11], deals with simple regions, i.e. connected regions that are topologically equivalent to closed disks. In this section, we show our consistency checking algorithm designed for connected regions can be adapted to cope with these two variants.

## 7.1 Cardinal Directions between Possibly Disconnected Regions

For two (possibly disconnected) regions $a, b$, similar to Dfn. 4, we write $\mathsf{dir}(a,b) = [d_{ij}]_{i,j=1}^{3}$ for the direction relation matrix of $a$ to $b$, where $d_{ij}$ is 1 if $a°\cap b_{ij} \neq \varnothing$, and 0 otherwise. A $3 \times 3$ Boolean matrix $M$ is valid if there exist two region $a, b$ such that $M = \mathsf{dir}(a,b)$. It is easy to see that all but the zero $3 \times 3$ Boolean matrices are valid. Each of these matrices represents a basic direction relation between possibly disconnected regions. We call the Boolean algebra generated by these JEPD relations the Cardinal Direction Calculus for possibly disconnected regions, denoted as $\text{CDC}_d$.

Consistency checking in $\text{CDC}_d$ is similar to that in CDC. Suppose $\mathcal{N} = \{v_i \delta_{ij} v_j\}_{i,j=1}^n$ is a network of basic $\text{CDC}_d$ constraints. Similar definitions of regular solutions, meet-free solutions, and canonical solutions can be defined in $\text{CDC}_d$. Moreover, suppose $\mathfrak{a}$ is a solution to $\mathcal{N}$. We can transform $\mathfrak{a}$ into a canonical solution $\mathfrak{a}' = \{a_i'\}_{i=1}^n$ of $\mathcal{N}$. That is, $\mathfrak{a}'$ is a regular, meet-free, and digital solution, and $\{I_x(a_i')\}_{i=1}^n$ and $\{I_y(a_i')\}_{i=1}^n$ are canonical sets of intervals. Write $m_i = I_x(a_i') \times I_y(a_i')$. We next show $\mathcal{N}$ has a maximal canonical solution. Actually, for each $i$, set $D_i$ to be the set of pixels in $m_i$ that are disallowed for violating some constraint in $\mathcal{N}$ (see Eq. 24), and set $b_i$ to be the region obtained by deleting all disallowed pixels from $m_i$ (see Eq. 25). We assert that $\mathfrak{b} = \{b_i\}_{i=1}^n$ is the maximal canonical solution of $\mathcal{N}$.

**Theorem 6.** *Suppose $\mathcal{N} = \{v_i \delta_{ij} v_j\}_{i,j=1}^n$ is a network of basic $\text{CDC}_d$ constraints. If $\mathcal{N}$ is consistent, then $\mathfrak{b} = \{b_i\}_{i=1}^n$ is the maximal canonical solution of $\mathcal{N}$, where $b_i$ is defined as in Eq. 25.*

*Proof.* The proof is similar to that for Lemma 18. □

This shows, to construct the maximal canonical solution of a network of basic $\text{CDC}_d$ constraints, we need not to compute the connected components of $b_i$.

We next adapt our main algorithm to determine the consistency of a network $\mathcal{N} = \{v_i \delta_{ij} v_j\}_{i,j=1}^n$ of basic $\text{CDC}_d$ constraints. As in the case of connected regions, Step 1 computes the projective IA networks $\mathcal{N}_x$ and $\mathcal{N}_y$. If either is inconsistent, then $\mathcal{N}$ is inconsistent. Otherwise, Step 2 constructs their canonical interval solutions, and Step 3 computes $b_i$ for each $i$. In case $\mathcal{M}(b_i) \neq m_i$ for some $i$ then $\mathcal{N}$ is inconsistent. Otherwise, we go to the next step. The above procedures, as in the connected case, need at most cubic time.

Since regions in a solution to $\text{CDC}_d$ constraints are allowed to be disconnected, we need not compute the connected components of each $b_i$. Therefore, we go directly to Step 5, where we need check if $\mathsf{dir}(b_i, b_j) = \delta_{ij}$ holds for each pair of $i, j$. Suppose $\delta_{ij}$ is represented as a 9-tuple $(d_{ij}^\phi)_{\phi=1}^9$. Similar to the connected case, we need only check whether $b_i^\circ \cap m_j^\phi$ is nonempty for each $\phi$ with $d_{ij}^\phi = 1$.

To check if $b_i^\circ \cap m_j^\phi$ is nonempty, it is sufficient to show that there is a pixel $p$ contained in $b_i \cap m_j^\phi$. Since $b_i$ may be disconnected, checking all boundary pixels is, however, insufficient. Note that checking all pixels in $m_i \cap m_j^\phi$ alone needs $O(n^2)$ time in the worst case for each pair $i, j$. This is undesirable since it will cost $O(n^4)$ time in total.

We next show this could also be simplified. For a digital region $b$ contained in $T$, write $B(b)$ for the Boolean matrix that represents $b$ (see Eq. 29), and define an $n_x \times n_y$ integer matrix $M(b)$ as

$$M(b)[k,l] = \sum \{B(b)[p,q] : 0 \leq p \leq k, 0 \leq q \leq l\} \qquad (47)$$

for each $0 \leq k < n_x$ and each $0 \leq l < n_y$. It is easily to see that $M(b)[k,l]$ is the number of pixels of $b$ which are also contained in the rectangle $[0, k+1] \times [0, l+1]$.

Given the Boolean matrix $B(b)$, $M(b)$ can be computed in $O(n^2)$ time by iteratively adding the $k$-th column to the $(k+1)$-th, and then iteratively adding the $p$-th row to the $(p+1)$-th.

**Lemma 25.** *Given a rectangle $r = [x^-, x^+] \times [y^-, y]$ and a digital region $b$, both contained in $T = [0, n_x] \times [0, n_y]$, $b$ contains a pixel in $r$ if and only if*
$$M(b)[x^- - 1, y^- - 1] + M(b)[x^+ - 1, y^+ - 1] >$$
$$M(b)[x^+ - 1, y^- - 1] + M(b)[x^- - 1, y^+ - 1], \qquad (48)$$
*where $M(r)[-1, l] = M(r)[k, -1] = 0$.*

*Proof.* Because $M(b)[k, l]$ denotes the number of pixels in $b$ which are also contained in $[0, k+1] \times [0, l+1]$, the number of pixels in $b$ which are contained in the rectangle $[x^-, x^+] \times [y^-, y^+]$ is $M(b)[x^+ - 1, y^+ - 1] - (M(b)[x^- - 1, y^+ - 1] + M(b)[x^+ - 1, y^- - 1]) + M(b)[x^- - 1, y^- - 1]$. The conclusion follows directly. □

Suppose $T \cap m_j^\phi = [x^-, x^+] \times [y^-, y^+]$. Since $b_i \subseteq T$, we have
$$b_i \cap m_j^\phi = b_i \cap (T \cap m_j^\phi) = b_i \cap [x^-, x^+] \times [y^-, y^+].$$

By the above lemma, we know $b_i \cap m_j^\phi$ contains a pixel if and only if Eq. 48 holds, which can be checked in constant time. So, given $M(b_i)$, the constraint $\delta_{ij}$ can be checked in constant time. Computing all $M(b_i)$ needs $O(n^3)$ time. Since there are $n^2$ constraints in total, Step 5, and thereby the who algorithm, can be finished in $O(n^3)$ time.

*Remark* 2. For a consistent $CDC_d$ network $\mathcal{N}$, the $O(n^5)$ algorithm proposed in [33] outputs a solution of $\mathcal{N}$ using possibly disconnected regions. It is difficult to extend this method to cope with cardinal directional constraints between connected regions. Actually, Navarrete *et al.* [23] tried to adapt this algorithm to cope with connected regions, and proposed an $O(n^4)$ algorithm for checking the consistency of a basic CDC network. The algorithm was based on Theorem 1 in [23], which actually implied the following proposition.

> For a basic CDC network $\mathcal{N}$, if all subnetworks of $\mathcal{N}$ involving four variables are consistent, then $\mathcal{N}$ is consistent.

However, the counterexample constructed in Section 3.4 of this paper shows this is not true.

## 7.2 Cardinal Directions between Simple Rgions

In the definition of direction relation matrix (Dfn. 4), we only assume connected regions. It is possible that these connected regions have holes. In the original work of Goyal and Egenhofer [11, 12], objects are represented as simple regions, *i.e.* regions that are topologically equivalent to closed disks. Interestingly, each direction relation matrix between connected regions can be realized by a pair of simple regions. Therefore, the set of cardinal direction relations between simple regions is the same as that between connected regions. We write $\text{CDC}_s$ for the qualitative calculus generated by these cardinal direction relations on the set of simple regions.

In this subsection, we show this difference between simple regions and connected regions does not affect the consistency of cardinal direction constraints. In particular, we prove that a consistent network of CDC constraints always has a solution only using simple regions. The idea is to transform each connected region in the maximal canonical solution into a simple region without changing the relations between these regions.

Suppose $a$ is a digital region that is connected. Clearly, $a$ has at most finite holes, where a *hole* of $a$ is the closure of a bounded connected component of the exterior of $a$. It is easy to see that for a digital region all holes are digital and simple regions.

**Definition 17** (contact points). Let $a$ be a digital region. A point $P = (k, l)$ is called a *contact point* of $a$ if

$$p_{k,l} \subseteq a \Leftrightarrow p_{k-1,l-1} \subseteq a \Leftrightarrow p_{k-1,l} \not\subseteq a \Leftrightarrow p_{k,l-1} \not\subseteq a. \tag{49}$$

Clearly, among the four pixels around a contact point, only two belong to $a$, and these two pixels are 8-neighbors, *i.e.* they are diagonally adjacent.

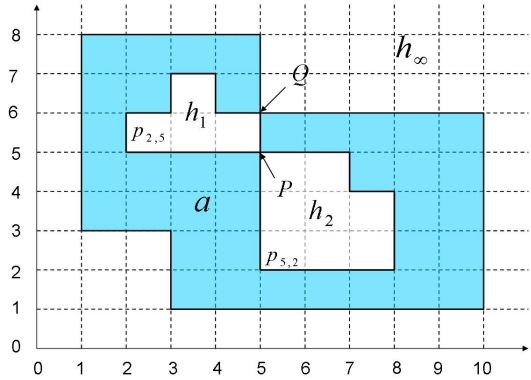

Figure 18: A digital region $a$ with two contact points $P = (5, 5)$ and $Q = (5, 6)$, where $h_1$ and $h_2$ are two holes of $a$, $h_\infty$ is the closure of the unbounded component of the exterior of $a$

Suppose $\mathfrak{a} = \{a_i\}_{i=1}^n$ is the maximal canonical solution of a basic CDC network. We next transform each $a_i$ into a simple region by deleting subpixels from $a$. Examples are show in Fig. 19 and Fig. 20.

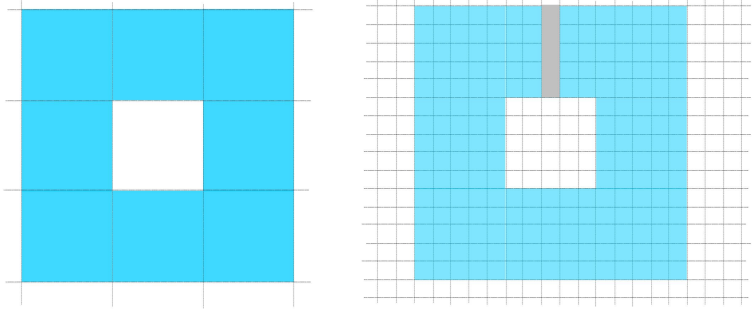

Figure 19: Transform a connected digital region without contact points into a simple region

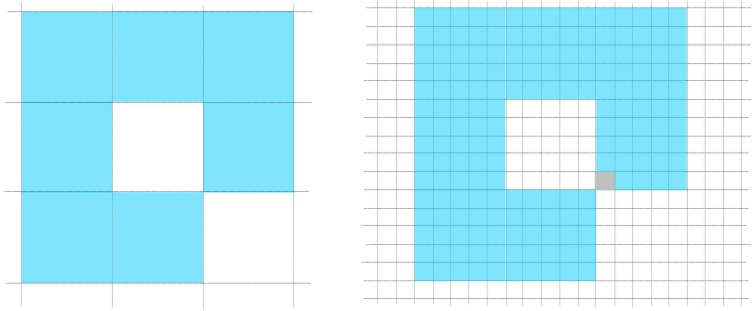

Figure 20: Transform a connected digital region with one contact points into a simple region

**Lemma 26.** *Suppose $\mathfrak{a} = \{a_i\}_{i=1}^n$ is the maximal canonical solution of a basic CDC network. Let $T$ be the frame of $\mathfrak{a}$, and let $m_j$ be the mbr of $a_j$ for each $j$. If $p_{kl}$ is a pixel which is contained in a hole of $a_i$, then there exists $j \neq i$ such that $p_{kl} \subseteq m_j$ and $a_i^\circ \cap m_j = \varnothing$.*

*Proof.* If $p_{kl}$ is contained in a hole if $a_i$, then there exist pixels contained in $a_i$ that are, respectively, above, below, right of, and left of $p_{kl}$. That is, there exist $k_1 < k < k_2$ and $l_1 < l < l_2$ such that $p_{k_1l}, p_{k_2l}$ and $p_{kl_1}, p_{kl_2}$ are all contained in $a_i$. By the construction of $a_i$, we know $p_{kl}$ is a pixel in $D_i$ (see Eq.24). This means, there exist $j \neq i$ and $\phi$ such that $a_i^\circ \cap m_j^\phi = \varnothing$ and $p_{kl} \subseteq m_j^\phi$. If $\phi \neq 5$, it is easy to see $m_j^\phi$ also contains at least one of the four 'guarding' pixels $p_{k_1l}, p_{k_2l}$ and $p_{kl_1}, p_{kl_2}$. This contradicts the fact that these pixels are contained in $a_i$. As a consequence, we know $\phi = 5$ and $m_j^\phi = m_j^5 = m_j$. This means, there is $j \neq i$ such that $p_{kl} \subseteq m_j$ and $a_i^\circ \cap m_j = \varnothing$. □

**Lemma 27.** *For two digital regions $a, b$, if $a' \subseteq a$ and $b' \subseteq b$ satisfy Eq. 50*

and, respectively, Eq. 51, then $\mathsf{dir}(a,b) = \mathsf{dir}(a',b')$.

$$\mathcal{M}(a) = \mathcal{M}(a') \text{ and for all pixels } p, \ p \subseteq a \text{ implies } (a')^\circ \cap p \neq \varnothing \quad (50)$$

$$\mathcal{M}(b) = \mathcal{M}(b') \text{ and for all pixels } p, \ p \subseteq b \text{ implies } (b')^\circ \cap p \neq \varnothing. \quad (51)$$

*Proof.* This follows directly from the definition of direction relation matrix. Note that for any $\phi = 1, \cdots, 9$ we have $a \cap m^\phi = \varnothing$ if and only if $a' \cap m^\phi = \varnothing$, where $m^\phi = b^\phi = (b')^\phi$. □

Recall a maximal canonical solution is a meet-free solution (cf. Dfn. 11). This implies that the mbrs of any two regions in a maximal canonical solution do not meet at a point. The following lemma is a consequence of this observation.

**Lemma 28.** *Suppose $\mathfrak{a} = \{a_i\}_{i=1}^n$ is the maximal canonical solution of a basic CDC network. If $P = (k,l)$ is a contact point of $a_1$, then among the four pixels around $P$, viz. $p_{k-1,l-1}, p_{kl}, p_{k-1,l}, p_{k,l-1}$, one is contained in a hole of $a_1$, two are contained in $a_1$, the other is contained in neither $a_1$ nor its holes.*

*Proof.* Without loss of generality, suppose $p_{kl}$ and $p_{k-1,l-1}$ are contained in $a_1$. If $p_{k-1,l}$ and $p_{k,l-1}$ are contained in neither $a_1$ nor its holes, then they are connected to a same pixel in the unbounded connected component of the exterior of $a_1$. This means $p_{kl}$ is separated from $p_{k-1,l-1}$, which contradicts the assumption that $a_1$ is a connected region. On the other hand, if each of $p_{k-1,l}$ and $p_{k,l-1}$ is contained in a hole of $a_1$, then by Lemma 26, $p_{k-1,l} \subseteq m_j$ and $p_{k,l-1} \subseteq m_{j'}$ for some $j, j'$ with $a_1^\circ \cap m_j = \varnothing$ and $a_1^\circ \cap m'_j = \varnothing$. Because $p_{kl}$ and $p_{k-1,l-1}$ are not contained in $m_j$ and $m'_j$, we know $m_j$ and $m'_j$ must meet at point $P$. This is impossible for maximal canonical solutions. Therefore, only one of $p_{k,l-1}$ and $p_{k-1,l}$ is contained in a hole of $a_1$. □

By the above lemma, there are four different types of contact points. See Fig. 3. For convenience, we denote each type as a 4-tuple of symbols taken from $\{h, a, x\}$. For the four pixels, we start from the top left corner and go clockwise. These four types are written, respectively, as $(haxa)$, $(ahax)$, $(xaha)$, and $(axah)$.

| h | a |   | a | h |   | x | a |   | a | x |
|---|---|---|---|---|---|---|---|---|---|---|
| a | x |   | x | a |   | a | h |   | h | a |

Table 3: Four types of contact points, where $h, a, x$ denote, respectively, a pixel that is contained in a hole of $a$, in $a$, and in neither.

A contact point can be removed by deleting a sub-pixel from the $a$-pixel which follows the $h$-pixel according to the sequence of the type of the contact point. Once all contact points are removed from each $a_i$, the remaining holes are quite easy to cope with.

**Theorem 7.** *Let $\mathcal{N} = \{v_i \delta_{ij} v_j\}_{i,j=1}^n$ be a network of basic CDC constraints. If $\mathcal{N}$ is consistent, then it has a solution $\mathfrak{a} = \{a_i\}_{i=1}^n$ such that each $a_i$ is a simple region.*

*Proof.* Suppose $\mathfrak{a} = \{a_i\}_{i=1}^n$ is the maximal canonical solution of $\mathcal{N}$. Then each $a_i$ is a connected digital region, which may have holes. We assert that, for each $i$, there exists a simple region $a_i'$ such that $a_i' \subseteq a_i$ and $\mathcal{M}(a_i') = \mathcal{M}(a_i)$, and $\mathsf{dir}(a_i, a_j) = \mathsf{dir}(a_i', a_j)$ and $\mathsf{dir}(a_j, a_i) = \mathsf{dir}(a_j, a_i')$ for any $j \neq i$.

To prove this statement, we first subdivide each pixel into 25 equal sub-pixels (see Fig. 21).

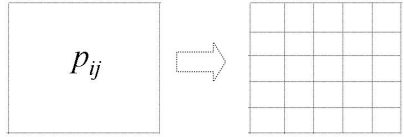

Figure 21: Subdivision of a pixel into 25 sub-pixels

For a contact point $P = (k, l)$ of $a_i$, without loss of generality, assume $P$ has type $(haxa)$. Removing the $1/25$ sub-pixel which contains $P$ from $a_i$, we obtain (after necessary regularization[1]) a new region (see Fig.22). This procedure can be applied to all contact points of $a_i$ at the same time. Write $a_i^*$ for the resulted region. Since each pixel $p$ of $a_i$ has at most four contact points, the revised region $a_i^*$ contains at least 21 of the 25 sub-pixels of $p$. This implies that $a_i^* \subseteq a_i$ and $\mathcal{M}(a_i^*) = \mathcal{M}(a_i)$. It is routine to check that $a_i^*$ is still connected but has fewer contact points as well as fewer holes. By Lemma 27 we know $\mathsf{dir}(a_i^*, a_j) = \mathsf{dir}(a_i, a_j)$ and $\mathsf{dir}(a_j, a_i^*) = \mathsf{dir}(a_j, a_i)$ for any $j \neq i$.

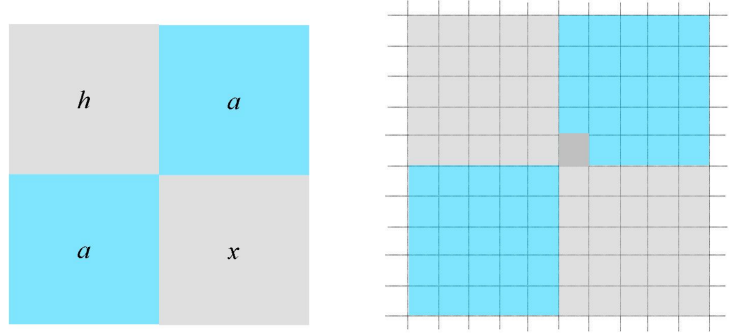

Figure 22: Remove a type $(haxa)$-contact point

Now, we have removed all contact points of $a_i$. This means, any two holes of $a_i^*$ are disjoint, and each hole of $a_i^*$ is disjoint from the closure of the unbounded component of the exterior of $a_i^*$. For each hole $h$ of $a_i^*$, select a pixel $p(h)$ in $h$ which has the highest $y$-index but whose left 4-neighbor is out of $h$. We cut a slot from $a_i^*$ to connect the hole $h$ with the exterior of $a_i^*$. For a pixel $p$ above $p(h)$, if $p$ and all pixels between $p$ and $p(h)$ are contained in $a_i^*$, then delete the middle column of sub-pixels from $p$ (see Fig. 19). After necessary regularization, we obtain another connected region which has fewer holes than $a_i^*$. Applying this operation to holes of $a_i^*$ one by one, we obtain a connected region $a_i'$, which has no holes. That is, $a_i'$ is a simple region. By Lemma 27, we know $\mathsf{dir}(a_i', a_j) = \mathsf{dir}(a_i, a_j)$ and $\mathsf{dir}(a_j, a_i') = \mathsf{dir}(a_j, a_i)$ for any $j \neq i$.

---

[1] Note here by regularization we mean the topological regularization of a set $a$, which is defined as $\overline{a^\circ}$.

We claim $\{a_i'\}_{i=1}^n$ is a solution of $\mathcal{N}$ which consists of simple regions. For any $i \neq j$, because $\mathcal{M}(a_j') = \mathcal{M}(a_j)$ and $\mathsf{dir}(a_i', a_j) = \mathsf{dir}(a_i, a_j)$, we know $\mathsf{dir}(a_i', a_j') = \mathsf{dir}(a_i, a_j)$. The conclusion holds already. □

As a conclusion, we know the satisfaction problem over $\mathrm{CDC}_s$ can be decided in the same way as that over CDC.

**Theorem 8.** *The consistency of a network of basic $CDC_s$ constraints can be determined in cubic time.*

# 8 Conclusion

This paper provided a cubic algorithm for checking consistency of basic CDC networks, which was earlier observed as impossible [23] for connected regions. If a basic CDC network is consistent, our algorithm also generates the maximal canonical solution. This general algorithm was then applied to solve two subproblems: the pairwise consistency problem and the weak composition problem. Through careful examination, we showed that the cardinal direction relation model used in [3] is not coherent, and showed by examples that the weak composition algorithm obtained in [32] is not always correct.

Although devised to solve cardinal directional constraints between connected regions, our main algorithm can also be adapted to cope with cardinal directional constraints between possibly disconnected regions as well as those between simple regions. For a basic network of $\mathrm{CDC}_d$ constraints, our algorithm determines in cubic time if it is consistent. Compared with the $O(n^5)$ algorithm reported in [33], our algorithm is more efficient. As for cardinal direction constraints over simple regions, we proved that a basic CDC network has a solution using only simple regions if it is consistent. This suggests that CDC does not distinguish between simply connectedness and connectedness.

Most potential applications of qualitative spatial reasoning require multiple aspects of space. Combining spatial constraints of different calculi is a very important problem in the research of qualitative spatial reasoning. Some work has been done in this direction (see *e.g.* [9, 16, 22]). In particular, [22] points out that reasoning with basic RCC8 and basic RA constraints is in P, but reasoning with basic RCC8 and basic $\mathrm{CDC}_d$ constraints is NP-Complete. Note that in RCC8, RA and $\mathrm{CDC}_d$, spatial variables are interpreted over possibly disconnected regions. It is still open if the above results for possibly disconnected regions still hold for connected regions. In particular, we do not know if reasoning with basic RCC8 and basic CDC constraints is still decidable if spatial variables are interpreted over connected regions. It is worth noting that connectedness is a tough requirement as far as topological relations are concerned. Although path-consistency suffices to determine if a basic RCC8 network is consistent, it was recently proved that deciding if a basic RCC8 network has a solution using connected regions is NP-Complete [30].

## Acknowledgements

This work was partially supported by NSFC (60673105, 60736011, 60621062), and a 973 Program (2007CB311003). The work of Sanjiang Li was also partially supported by a Microsoft Research Professorship.

## Appendix A: Proof of Lemma 6

**Lemma 6.** *Suppose $\delta = [d_{ij}]_{i,j=1}^{3}$ is a basic CDC relation. Then the x-projective interval relation $\iota^x(\delta)$ is the IA relation associated to the vector $(d_1, d_2, d_3)$, i.e. $(I, J) \in \iota^x(\delta)$ iff $\mathsf{dir}(I, J) = (d_1, d_2, d_3)$, where $d_j$ is 0 if $\Sigma_{i=1}^{3} d_{ij} = 0$ and 1 otherwise.*

*Proof.* We need to prove two things. First, suppose $(a, b) \in \delta$. We need show $\mathbf{v}(I_x(a), I_x(b))$, the direction relation vector of $(I_x(a), I_x(b))$ is $(d_1, d_2, d_3)$. Second, suppose $\mathbf{v}(I_1, I_2) = (d_1, d_2, d_3)$. We need show there exist two connected regions $a, b$ such that $(a, b) \in \delta$ and $I_1 = I_x(a), I_2 = I_x(b)$.

The first part can be proved as follows. Set $I_x(a) = [x^-(a), x^+(a)]$, $I_x(b) = [x^-(b), x^+(b)]$. By the definition of cardinal direction matrix, $x^-(a) \geq x^-(b)$ if and only if $a^\circ \cap b_{i1} = \varnothing$, i.e. $\delta_{i1} = 0$, for $i = 1, 2, 3$; $(x^-(a), x^+(a)) \cap (x^-(b), x^+(b)) = \varnothing$ if and only if $a^\circ \cap b_{i2} = \varnothing$, i.e. $\delta_{i2} = 0$, for $i = 1, 2, 3$; $x^+(a) \leq x^+(b)$ if and only if $a^\circ \cap b_{i3} = \varnothing$, i.e. $\delta_{i3} = 0$, for $i = 1, 2, 3$. Therefore, the vector $(d_1, d_2, d_3)$ as defined here is the direction relation vector of $(I_x(a), I_x(b))$. In other words, $(I_x(a), I_x(b))$ is an instance of $\iota_x(\delta)$.

To prove the second part, we note that $\delta$ as a direction relation matrix is 4-connected (see Prop. 2). The proof is by construction. Take

$$\delta = \begin{pmatrix} 1 & 1 & 0 \\ 0 & 1 & 0 \\ 1 & 1 & 0 \end{pmatrix}$$

as example. In this case $(d_1, d_2, d_3) = (1, 1, 0)$. Suppose $\mathsf{dir}(I_1, I_2) = (d_1, d_2, d_3)$. Set $I_1 = [u, v]$, $I_2 = [s, t]$. Then we have $u < s < v \leq t$. There are two subcases. If $v < t$, put $I_1$ and $I_2$ on the $x$-axis of the orthogonal basis of the plane. Set $a, b$ to be the polygons in Fig. 23 (left). It is clear $(a, b) \in \delta$ hold. The case of $v = t$ is similar and illustrated in Fig. 23 (right). The same method applies to all the other basic CDC relations. □

# Appendix B: Proof of Lemma 9

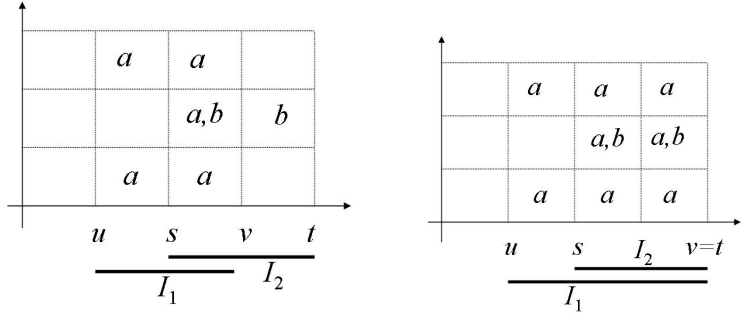

Figure 23: Construction of Cardinal Direction Relations from $x$-projective relations

**Lemma 9.** *An IA network $\mathcal{N} = \{v_i \iota_{ij} v_j\}_{i,j=1}^n$ of constraints over $\mathcal{B}_{int}^*$ is consistent if and only if $\widehat{\mathcal{N}} = \{v_i \widehat{\iota}_{ij} v_j\}_{i,j=1}^n$ is satisfiable.*

*Proof.* We first observe that $\widehat{\mathcal{N}}$ is a refinement of $\mathcal{N}$. Therefore, if $\widehat{\mathcal{N}}$ is consistent, so is $\mathcal{N}$. On the other hand, suppose $\{I_i = [u_i^-, u_i^+]\}_{i=1}^n$ is a solution to $\mathcal{N}$. We construct a solution to $\widehat{\mathcal{N}}$.

We call a point $u \in M = \{u_i^-, u_i^+\}_{i=1}^n$ a meet point if there exist $i \neq j$ such that $u_i^+ = u_j^-$, i.e. $I_i$ meets $I_j$. We use induction on the number $K$ of meet points. If there is no meet point, $\{I_i = [u_i^-, u_i^+]\}_{i=1}^n$ is also a solution to $\widehat{\mathcal{N}}$. Suppose the result holds for $K = m$. We show this also holds for $K = m + 1$. Suppose $w = u_{i_0}^-$ is the largest meet point. We define a new solution $\{J_i = [v_i^-, v_i^+]\}_{i=1}^n$ to $\mathcal{N}$ which has fewer meet points. Set

$$v_i^+ = u_i^+ \quad \text{and} \quad v_i^- = \begin{cases} w + \varepsilon/4, & \text{if } u_i^- = w, \\ u_i^-, & \text{otherwise.} \end{cases}$$

where $\varepsilon$ is the smallest distance between different points in $M$. We next show each pair $(J_i, J_k)$ is also an instance of $\iota_{ik}$. To this end, we need only consider the case when either $J_i \neq I_i$ or $J_k \neq I_k$. Note that $J_i \neq I_i$ if and only if $u_i^- = w$. This means we need only consider the cases when $u_i^- = w$ or $u_k^- = w$. Recall $v_i^+ = u_i^+$, $v_k^+ = u_k^+$, and $v_i^- \in \{u_i^-, w + \varepsilon/4\}$, $v_k^- \in \{u_k^-, w + \varepsilon/4\}$.

Case 1. If $u_i^- = w$ and $u_k^- = w$, then $v_i^- = v_k^- = w + \varepsilon/4$. The ordering of $v_i^-, v_i^+, v_k^-, v_k^+$ is the same as that of $u_i^-, u_i^+, u_k^-, u_k^+$.

Case 2. If $u_i^- = w$ and $u_k^+ = w$, then $u_k^- < u_k^+ = u_i^- < v_i^+$ and $v_k^- < v_k^+ < v_i^- < v_i^+$ and $v_k^- = u_k^-$. This means $I_k \mathsf{m} I_i$ and $J_k \mathsf{p} J_i$.

Case 3. If $u_k^- = w$ and $u_i^+ = w$, then, similar to Case 2, we have $I_i \mathsf{m} I_k$ and $J_i \mathsf{p} J_k$.

Case 4. If $u_i^- = w$ and $u_k^+ \neq w$, the ordering of $v_i^-, v_i^+, v_k^-, v_k^+$ is the same as that of $u_i^-, u_i^+, u_k^-, u_k^+$.

Case 5. If $u_k^- = w$ and $u_i^+ \neq w$, the ordering of $v_i^-, v_i^+, v_k^-, v_k^+$ is the same as that of

So $\{J_i\}_{i=1}^n$ is also a solution to $\mathcal{N}$. Note that either Case 2 or Case 3 is true, because $w$ is a meet point. Moreover, $w$ is not a meet point in the new solution anymore since all 'meets' instances at this point has been changed to 'before' instance. Therefore, $\{J_i\}_{i=1}^n$ has fewer meet points than $\{I_i\}_{i=1}^n$. By the induction hypothesis, we know $\widehat{\mathcal{N}}$ has a solution. This ends the proof. □